\documentclass{article}
\pdfoutput=1

\usepackage[preprint]{neurips_2023}

\usepackage[utf8]{inputenc} %
\usepackage[T1]{fontenc}    %
\usepackage{hyperref}       %
\usepackage{url}            %
\usepackage{booktabs}       %
\usepackage{amsfonts}       %
\usepackage{nicefrac}       %
\usepackage{microtype}      %
\usepackage{xcolor}         %

\usepackage[small]{caption}
\usepackage{graphicx}
\usepackage{amsmath}
\usepackage{amsthm}
\usepackage{algorithm}
\usepackage{algorithmic}
\usepackage{wrapfig}

\usepackage[scientific-notation=true]{siunitx}

\usepackage{marginnote}

\usepackage[textsize=tiny, textwidth=1.9in, colorinlistoftodos]{todonotes}

\usepackage[normalem]{ulem}

\newlength\myindent
\usepackage{subcaption}
\usepackage{enumitem}

\usepackage{multirow}
\usepackage{makecell}

\usepackage{pgfplots}
\usepackage{pgfplotstable}
\usepgfplotslibrary{fillbetween}
\pgfplotsset{
  compat=1.18,
  MediumBarPlot/.style={
    font=\small,
    ybar,
    width=\linewidth,
    ymin=0,
    xtick=data,
    xticklabel style={text width=2.5cm, align=center},
    xtick pos=left,
  },
  BlueBars/.style={
    fill=blue!20, bar width=2.5pt
  },
  PurpleBars/.style={
    fill=purple!70, bar width=2pt
  },
  GreenBars/.style={
    fill=ForestGreen, bar width=3.5pt
  },
  YellowBars/.style={
    fill=Goldenrod!80, bar width=3.5pt
  },
  TealBars/.style={
    fill=teal!80, bar width=2pt
  },
  BrownBars/.style={
    fill=brown!80, bar width=2.5pt
  },
}

\newcommand{\groupSet}[0]{\ensuremath{\mathcal{\textbf{G}}}}
\newcommand{\taskSet}[0]{\ensuremath{\mathcal{T}}}
\newcommand{\loss}[0]{\ensuremath{\mathcal{L}}}

\DeclareMathOperator*{\argmin}{argmin}

\newcommand{\score}[0]{\ensuremath{\mathcal{S}}}
\newcommand{\arch}[0]{\ensuremath{\mathcal{A}}}

\newcommand{\hyperParam}[0]{\ensuremath{\mathcal{HP}}}
\newcommand{\probabilityDistribution}[0]{\ensuremath{\mathcal{P}}}
\newcommand{\neuron}[0]{\ensuremath{\mathcal{N}}}
\newcommand{\layerNN}[0]{\ensuremath{\mathcal{L}}}
\newcommand{\featureNN}[0]{\ensuremath{\mathcal{F}}}

\usepackage[capitalise,noabbrev]{cleveref}

\title{Task Grouping for Automated Multi-Task Machine Learning via Task Affinity Prediction}
\author{
Afiya Ayman\\
Pennsylvania State University
\And
Ayan Mukhopadhyay\\
Vanderbilt University
\And
Aron Laszka\\
Pennsylvania State University}

\begin{document}

\setcounter{figure}{0}
\setcounter{table}{0}
\setcounter{algorithm}{0}

\setlength{\marginparwidth}{1.5in}
\setlength{\marginparsep}{0.325in}

\maketitle

\begin{abstract}
When a number of similar tasks have to be learned simultaneously,
multi-task learning (MTL) models can attain significantly higher accuracy than single-task learning (STL) models.
However, the advantage of MTL depends on various factors, such as the similarity of the tasks, the sizes of the datasets, and so on; in fact, some tasks might not benefit from MTL and may even incur a loss of accuracy compared to STL.
Hence, the question arises: which tasks should be learned together?
Domain experts can attempt to group tasks together following intuition, experience, and best practices, but manual grouping can be labor-intensive and far from optimal.
In this paper, we propose a novel automated approach for task grouping.
First, we study the affinity of tasks for MTL using four benchmark datasets that have been used extensively in the MTL literature, focusing on neural network-based MTL models.
We identify inherent task features and STL characteristics that can help us to predict whether a group of tasks should be learned together using MTL or if they should be learned independently using STL.
Building on this predictor, we introduce a randomized search algorithm, which employs the predictor to minimize the number of MTL trainings performed during the search for task groups.
We demonstrate on the four benchmark datasets that our predictor-driven search approach can find better task groupings than existing baseline approaches.
\end{abstract}
\section{Introduction}
\label{sec:intro}

Multi-task learning (MTL) is an approach for learning more than one machine learning task in parallel to improve generalization performance~\citep{caruana1997multitask}. Learning multiple tasks together can also reduce inference time, which is particularly useful in estimating different attributes in visual inputs~\citep{standley2020tasks}. MTL improves the accuracy of individual tasks by utilizing related information in other tasks through a process known as \textit{inductive transfer}, i.e., the learning architecture is designed to introduce inductive bias to favor hypothesis classes that perform well across the set of tasks~\citep{caruana1997multitask}. 
MTL has proven to be effective in multiple domains, including computer vision (e.g., \citep{zhang2014facial,liu2015multi,lee2016asymmetric}), health informatics (e.g., \citep{puniyani2010multi,zhou2011multi}), and natural language processing (e.g., \citep{luong2015multi,wu2015collaborative}). However, simply learning a set of tasks together does not guarantee improved performance; in fact, performance can degrade compared to training individual models for each task, a phenomenon known as \textit{negative transfer}~\citep{standley2020tasks}. This observation raises two critical questions: \textit{Which tasks can be learned together to improve generalization for each task? And more importantly, how can we identify such groups of tasks?}

While MTL has been used extensively across application domains, the problem of designing principled approaches for identifying tasks that should be trained together has received significantly less attention. In general, tasks related to each other tend to benefit from %
MTL; the question of identifying tasks that should be trained together, hence, reduces to defining and identifying \textit{relatedness} between the tasks. Unfortunately, these relationships can be complex and abstract; for example, perhaps non-intuitively, transfer relationships (based on transfer learning) among tasks are not particularly good predictors of multi-task relationships~\citep{standley2020tasks}. One approach for identifying related tasks is leveraging human expertise~\citep{zhang2021survey}. However, human experts often rely on intuition rather than complex abstractions learned through data; also, possible combinations grow exponentially as the number of tasks increases, thereby rendering this manual approach infeasible in practice. 

There have been systematic efforts to open the black box of MTL and investigate which tasks should be learned together (and why). \citet{ben2008notion}
examine the similarities between the underlying sample-generating distributions across tasks and provide provable guarantees for MTL. \citet{bingel2017identifying}
provide an in-depth analysis of task interactions for the single domain of natural language processing but do not provide an algorithmic approach for selecting tasks for MTL. \citet{standley2020tasks} and \citet{fifty2021efficiently} propose approaches to identify groupings of tasks to improve accuracy given a limited computational budget for inference; however, their approaches consider only a small number of tasks. Motivated by prior work~\citep{bingel2017identifying,standley2020tasks,li2017better}, we provide a significantly deeper insight into attributes that identify the relatedness of tasks for MTL and provide a computationally efficient algorithm for grouping tasks.

Specifically, we study task relationships using four benchmark datasets that have been widely used in MTL. We explore pairwise and groupwise affinity among tasks, where affinity refers to a general property of task relatedness pertaining to MTL. We use relative MTL gain (defined as the relative gain of MTL over STL in terms of reducing prediction error)  to measure affinity. Then, we train a task-affinity predictor and show that it can be combined with a randomized search algorithm to design a computationally efficient algorithm for grouping tasks for MTL.  We provide a brief overview of prior work in \cref{sec:related}. Then, we describe the benchmark datasets and our experimental setup in \cref{sec:experiment_setup}. We study task affinity for MTL in \cref{sec:affinity}. We present our algorithmic approach for efficient task grouping in \cref{sec:approach} and experimental results in \cref{sec:result}.
\section{Prior Work on Task Relationships}
\label{sec:related}

Several recent studies have aimed to discover the underlying structure of tasks for transfer learning \citep{pal2019zero,dwivedi2019representation,achille2021information,zhuang2020comprehensive}.
\citet{zhang2021survey}
and
\citet{ruder2017overview}
provide detailed surveys that include various established 
methods to characterize the relatedness of tasks, such as feature learning, low-rank, task clustering, task relation learning, and decomposition approaches for MTL.
However, \citet{standley2020tasks}
show that transfer-learning algorithms often do not carry over task similarity to the multi-task learning domain, and propose a task-grouping framework that assigns tasks to networks to achieve the best overall prediction accuracy.
However, their framework can be prohibitively expensive computationally since it exhaustively searches the exponential space of possible groupings. \citet{bingel2017identifying}
present a systematic study of task-relatedness for pairs of common NLP tasks. \citet{shiri2022highly}
group tasks into clusters using a shared feature extractor across all tasks.

Recently, \citet{fifty2021efficiently} also propose a task-grouping framework where they train one single network with all tasks but track pairwise task affinities during training, which are then used to approximate higher-order affinities. While we take a similar approach, we explore significantly richer abstractions to combine the pairwise affinities. Moreover, prior work by \citet{fifty2021efficiently} applies to significantly smaller settings, requiring training one extensive network with all tasks and explicitly measuring pairwise affinity between all these tasks. In contrast, we present a scalable pairwise-affinity predictor, which uses features that can be computed easily. Further, the grouping search used by \citet{fifty2021efficiently} is a simple exhaustive search, which---as we show later---is infeasible for the benchmark datasets (e.g., 9 tasks in the setting of \citet{fifty2021efficiently} vs.\ 139 tasks in a benchmark in our analysis).

\textbf{Comparison to Prior Work:}
Our contribution in this paper is two-fold. First, we provide a study of the many factors that help in the understanding of task-relatedness, and based on these factors, we introduce task-affinity predictors that can estimate the relative gain of MTL over STL for a group of tasks (in terms of reducing loss), providing more accurate predictions than prior work.
Second, we show how the predictors can be combined with a randomized search algorithm to obtain efficient groupings for MTL, without exhaustively exploring the search space of possible task-groupings.

Our work is most closely related to that of \citet{standley2020tasks} with a critical difference: we train multi-task models for only a small number of groups from the space of all possible task groups. This difference is critical; \citet{standley2020tasks}
explore settings with two sets of 5 tasks for MTL, whereas we consider benchmarks with significantly more tasks (one of our %
datasets have 139 tasks for MTL). 

\section{Experimental Setup}
\label{sec:experiment_setup}

The datasets and experimental framework that we use for STL and MTL are critical to describing the rest of our analysis. Hence, we begin by describing our datasets and experimental setup first.  

\subsection{Datasets}
We study task relationships and demonstrate our approach using four benchmark datasets, which have been widely used in multi-task learning studies~\citep{zhang2021survey}. 
The characteristics of these datasets are summarized in \cref{tab:datadesc} (\cref{appendix:dataset_desc}).

\begin{itemize}[noitemsep, topsep=0pt,leftmargin=*]
    \item \textbf{School}~\citep{bakker2003task}: The goal is to predict exam scores for students from 139 schools given both school and student-specific attributes. Each school is a separate task.
    
    \item \textbf{Chemical}, a.k.a.\ MHC-I~\citep{jacob2008clustered}: The goal is to predict whether a peptide binds a molecule. The dataset contains binding affinities of 35 different molecules, i.e., 35 tasks.
    
    \item \textbf{Landmine}~\citep{xue2007multi}: This dataset consists of information collected from 29 landmine fields, each treated as a task, and the goal is to predict whether a data point is a landmine. 
    
    \item \textbf{Parkinson}~\citep{jawanpuria2015efficient}: This dataset consists of voice recordings for 42 patients, each treated as a task. The goal is to predict the disease (Parkinson's) symptom scores for the patients.
\end{itemize}

\subsection{STL and MTL Architectures}
In our experiments with single-task %
and multi-task learning %
on the four benchmark datasets, we employ feed-forward neural network (NN) models, Support Vector Machines (SVM), and eXtreme Gradient Boosting (XGBoost) trees (details in \cref{appendix:mtl_architectures}). The main text presents the results of NN-based STL and MTL experiments, given their numerous advantages, such as performance and broad applicability. Due to lack of space, the results of experiments with SVM and XGBoost models are provided in the appendix. 
Note that the specific MTL architectures are not a vital component to our contribution. 
In fact, \emph{our goal is orthogonal to the performance of specific MTL methods, and our task grouping approach can be applied to and potentially benefit a wide range of MTL methods.}

\textbf{NN Hyper-Parameter Search:} We conduct a neural architecture search on randomly chosen task groups from each of the four benchmark datasets to select the MTL architectures. We initialize the randomized architecture search with separate task-specific layers and some shared layers among all the tasks. We explore variations in the number of hidden layers, number of neurons per layer, and learning rate to minimize overall loss across the tasks and model complexity. The best architecture discovered during the search for each benchmark is adopted as the final MTL architecture.
    
\textbf{Implementation and Training:} We implement neural network models using Keras with TensorFlow \cite{abadi2016tensorflow}, minimizing either mean squared error (School and Parkinson) or binary cross-entropy (Chemical and Landmine) with the Adam optimizer \cite{kingma2014adam}. We train using an Intel Xeon G6252 CPU with an Nvidia V100 GPU and 128GB RAM, utilizing 20 cores.
We provide all software implementation for this paper as part of the supplementary material.

\section{Study of Task Affinity}
\label{sec:affinity}

We begin by studying relationships between different tasks. Our approach is motivated by prior work by \citet{zamir2018taskonomy}
(who evaluate such relationships in the context of transfer learning), \citet{bingel2017identifying}, \citet{standley2020tasks}, \citet{fifty2021efficiently}, and \citet{shiri2022highly}
. However, we explore a much wider array of features to gain a significantly deeper understanding of how relationships between tasks influence the performance of multi-task learning. 

\subsection{Pairwise Multi-Task Learning Affinity}
\label{subsec:pairwise_affinity}

Our first goal is to understand whether \textit{pairwise task affinity} can be predicted from the relationship between two tasks $t_i$ and $t_j$. By \textit{affinity}, we mean the general property of relatedness between tasks pertaining to MTL. We measure affinity as the \emph{relative MTL gain}, which is the improvement in predictive performance when tasks $t_i$ and $t_j$ are trained together with MTL instead of individually with STL, i.e., $\frac{\sum \textit{STL loss} - \textit{MTL loss} }{\sum \textit{STL loss}}$.

We predict affinity between a pair of tasks using a set of features. These features can be intrinsic to an individual task, i.e., capture the characteristics of the training sample of the task (e.g.,  standard deviation in the target attribute), or capture properties of STL applied to the individual task (e.g., learning curve). We refer to such features as \textit{single-task features}. We use a second set of features to capture the relationship between the single-task features of a pair of tasks and their training samples. We refer to such features as \textit{pairwise-task features}. Below, we provide a high-level overview of both types of features; we describe them in detail in \cref{appendix:feature_desc}. 

\textbf{Single-Task Features:} In addition to the features that \citet{bingel2017identifying} study for pairwise MTL in the NLP domain, we explore more features intrinsic to individual tasks,
 such as the variance ($\sigma^2$) and
standard deviation ($\sigma$)  of the target attribute and the average distances among the training samples of each task. Similar to \citet{bingel2017identifying}, we also consider gradients of the single-task learning curves at various stages of training. We also fit a logarithmic curve $y = a \cdot log(x) + b$ to the learning curve, where $y$ is the relative change in prediction loss and $x$ is the time steps in training, and use the parameters $a$ and $b$ as single-task features.

\textbf{Pairwise-Task Features:} These features describe the relationship between two tasks, $t_i$ and $t_j$. The features include the difference in the sample sizes of $t_i$ and $t_j$, the variance ($\sigma^2$) and the standard deviation ($\sigma$) of the target attribute when the training samples of $t_i$ and $t_j$ are combined, and the average distance between the training samples of $t_i$ and $t_j$, among others.
For distance-related features, we compute the distance \textit{between} the samples of $t_i$ and $t_j$, denoted by $t_i \leftrightarrow t_j$, and the distance \textit{among} the combined samples $t_i$ and $t_j$, denoted by $t_i+ t_j$.  
Depending on the data type, we compute distances using different measures, i.e., we use Euclidean ($d_E$) and Manhattan ($d_M$) distances for all benchmarks but additionally, we consider Hamming distance ($d_H$) for Chemical data. 
In the School benchmark, some numerical attributes have vast differences between the ranges of their values. To avoid biasing the results due to these differences (attributes with higher magnitude values being considered more important), we normalize the values of each attribute.
We normalize the computed pairwise-task features in multiple ways, i.e., by product and sum.
For example, we normalize the average Euclidean distance between the samples of two tasks\vphantom{$d_{E_{t_i}}$}, $\smash[b]{d_{E_{(t_i\leftrightarrow t_j)}}}$, 
with the sum of the average distances among the samples of each task, $d_{E_{t_i}}$ and $d_{E_{t_j}}$. 
We also consider the difference between the learning curve gradients of $t_i$ and $t_j$ from their respective STL curves. 

Similar to \citet{fifty2021efficiently}, we also train all tasks in a hard parameter sharing MTL model and measure how much one task’s gradient update affects the other tasks' loss. To stay consistent with \citet{fifty2021efficiently}, we use their formulation to compute this property (more details in \cref{appendix:feature_desc}) referring to this as \textit{inter-task affinity} (denoted by $\mathcal{Z}_{t_i \leftrightarrow t_j}$). We can use these affinity measures as another pairwise-task feature.  
From the same MTL model, we also collect the network parameters, $\mathcal{W}$, associated with task-specific
output layers. 
Specifically, we compose a vector consisting of the weights and biases $\mathcal{W}_{t}$ of the task-specific output layer of each task $t$. According to \cite{shiri2022highly}, given an identical network architecture, the manner in which these task-specific vectors are used in the final prediction can reveal task relationships. Hence, we use the dot products of these task-specific weight vectors, $\mathcal{W}_{t_i}\cdot\mathcal{W}_{t_j}$, as another pairwise-task feature for tasks $t_i$ and $t_j$. %

We have a total of 70 features for the School, Landmine, and Parkinson benchmarks and 64 features for the Chemical benchmark, which describe the tasks and their pairwise relationships. Among these, 34 features represent information about an individual task $t$, and the remaining features capture the pairwise relationship between two tasks $t_i$ and $ t_j$.

\pgfplotstableread[col sep=comma]{data/Pairwise_Individual_Usefulness_School_avg_NN.csv}\nnPairwiseSch
\pgfplotstableread[col sep=comma]{data/Pairwise_Individual_Usefulness_Chemical_avg_NN.csv}\nnPairwiseChem
\pgfplotstableread[col sep=comma]{data/Pairwise_Individual_Usefulness_Landmine_avg_NN.csv}\nnPairwiseLM
\pgfplotstableread[col sep=comma]{data/Pairwise_Individual_Usefulness_Parkinsons_avg_NN.csv}\nnPairwisePK

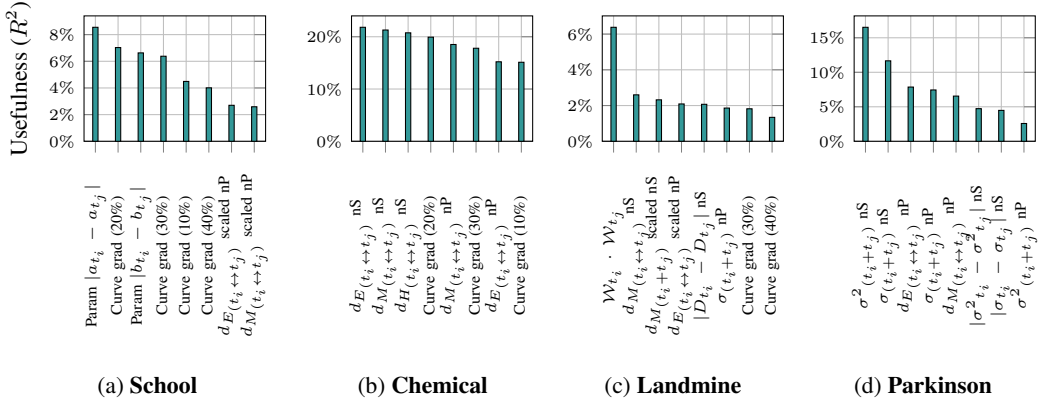
\begin{figure}[t]
    \begin{subfigure}[b]{0.286\linewidth}
    \begin{tikzpicture}[]
            \begin{axis}
            [
                MediumBarPlot,
                width = \linewidth,
                height = 3.25cm,
                ylabel = Usefulness (\textit{$R^2$}),
                label style = {align = center, font = \footnotesize},
                grid=major,
                font=\tiny,
                yticklabel=\pgfmathprintnumber{\tick}{$\%$},
                xtick = data,
                xmin = -0.5, xmax = 7.5,
                xticklabels from table={\nnPairwiseSch}{Feature},
              x tick label style={font = \tiny,rotate = 90, anchor=east},
            ]
            \addplot [TealBars] table [y={Avg_R_SQUARE}, x expr=\coordindex] {\nnPairwiseSch};
            \end{axis}
        \end{tikzpicture}
        \caption{\textbf{School}}
        \label{subfig:NNpair_sch}
    \end{subfigure}%
    \begin{subfigure}[b]{0.238\linewidth}
        \begin{tikzpicture}[]
                \begin{axis}
                [
                    MediumBarPlot,
                    width = 1.2\linewidth,
                    height = 3.25cm,
                    label style = {align = center, font = \footnotesize},
                    grid=major,
            yticklabel=\pgfmathprintnumber{\tick}{$\%$},
                    xtick = data,
                    xmin = -0.5, xmax = 7.5,
                     font=\tiny,
                    xticklabels from table={\nnPairwiseChem}{Feature},
                  x tick label style={font = \tiny,rotate = 90, anchor=east},
                ]
                \addplot [TealBars] table [y={Avg_R_SQUARE}, x expr=\coordindex] {\nnPairwiseChem};
                \end{axis}
            \end{tikzpicture}
            \caption{\textbf{Chemical}}
            \label{subfig:NNpair_chem}
        \end{subfigure}%
        \begin{subfigure}[b]{0.238\linewidth}
        \centering
        \begin{tikzpicture}[]
                \begin{axis}
                [
                    MediumBarPlot,
                    width = 1.2\linewidth,
                    height = 3.25cm,
                    label style = {align = center, font = \footnotesize},
                    grid=major,
                     font=\tiny,
                    yticklabel=\pgfmathprintnumber{\tick}{$\%$},
                    xtick = data,
                    xmin = -0.5, xmax = 7.5,
                    xticklabels from table={\nnPairwiseLM}{Feature},
                  x tick label style={font = \tiny,rotate = 90, anchor=east},
                ]
                \addplot [TealBars] table [y={Avg_R_SQUARE}, x expr=\coordindex] {\nnPairwiseLM};               
                \end{axis}
            \end{tikzpicture}
            \caption{\textbf{Landmine}} 
            \label{subfig:NNpair_lm}
        \end{subfigure}%
        \begin{subfigure}[b]{0.238\linewidth}
        \centering
        \begin{tikzpicture}[]
                \begin{axis}
                [
                    MediumBarPlot,
                    width = 1.2\linewidth,
                    height = 3.25cm,
                     font=\tiny,
                    label style = {align = center, font = \footnotesize},
                    grid=major,
                    yticklabel=\pgfmathprintnumber{\tick}{$\%$},
                    xtick = data,
                    xmin = -0.5, xmax = 7.5,
                    xticklabels from table={\nnPairwisePK}{Feature},
                  x tick label style={font = \tiny,rotate = 90, anchor=east},
                ]
                \addplot [TealBars] table [y={Avg_R_SQUARE}, x expr=\coordindex] {\nnPairwisePK};
                \end{axis}
            \end{tikzpicture}
            \caption{\textbf{Parkinson}}
            \label{subfig:NNpair_pk}
        \end{subfigure}
\caption{Most useful features for predicting relative MTL gain when two tasks are trained together. We describe the features in \cref{appendix:feature_desc}. The suffixes `nP' and `nS' stand for normalized by product and normalized by sum, respectively. The most useful features are related to distances between samples,  STL curve gradients, and variance and standard deviation of the target attributes.}%
\vspace{-1em}
\label{fig:predictor_pairwise}
\end{figure}

\textbf{Experiments:}
We first train single-task neural networks for all the tasks in all four benchmarks. We also train a pairwise multi-task model for each pair of tasks, yielding $9,591$ pairs for School (number of tasks, $n=139$), $595$ pairs for Chemical ($n=35$), $406$ pairs for Landmine ($n=29$), and $861$ pairs for the Parkinson ($n=42$) dataset.
The average loss over all tasks with STL and MTL demonstrate the usefulness of MTL (presented in \cref{tab:comparison_with_SOTA_MTL}). 
To investigate which task-relation features help in predicting affinity between a pair of tasks, we measure the Pearson correlation coefficient between each feature and the relative performance gain due to MTL. %
We perform an individual hyper\-parameter search to build a pairwise MTL-gain predictor for each pairwise feature (\cref{appendix:NAS}) and measure the feature's predictive performance ($R^2$) using the best-performing architecture
(\cref{fig:predictor_pairwise}). 
Finally, we perform a hyperparameter and feature-selection search to build a pairwise MTL-gain predictor.

\textbf{Findings:}
\cref{fig:predictor_pairwise} shows the 8 most useful pairwise-task features for predicting relative MTL gain for pairs of tasks, with the vertical bars indicating the variance explained ($R^2$ value) by each feature (usefulness of all features are presented in \cref{appendix:Predictor_Usefulness_TR}). Due to space constraints, we shorten the names of some features (full names with descriptions in  \cref{appendix:feature_desc}). 
We present the Pearson correlation of each feature with relative MTL gain in 
\cref{appendix:PearsonCorr_Pairwise}.

We find that various measures of distance between the samples of tasks ($t_i\leftrightarrow t_j$) and among the combined samples of tasks ($t_i+t_j$) are useful for predicting MTL gains on all four datasets (\cref{fig:predictor_pairwise}). 
We also find some single-task features, such as variance and standard deviation of the target attributes,
to be helpful for predicting MTL gain for Landmine and Parkinson datasets (\cref{subfig:NNpair_lm,subfig:NNpair_pk}).
Features describing the STL curves (curve gradients at different stages and fitted parameters) appear to be highly correlated and useful for predicting pairwise MTL gain. 
We do not find \textit{inter-task affinity}, ($\mathcal{Z}_{t_i \leftrightarrow t_j}$) \citep{fifty2021efficiently} to be generally useful for predicting \textit{relative MTL gain} despite having a high correlation on 2 out of 4 benchmarks (\cref{appendix:PearsonCorr_Pairwise,appendix:Predictor_Usefulness_TR}).
Pairwise-task features, such as distance metrics,  variance and standard deviation of target attribute, inter-task affinity, and learning-curve gradient features, exhibit strong correlations with the relative MTL gain on each benchmark (\cref{fig:pearsonCorr_pair} in \cref{appendix:PearsonCorr_Pairwise}).

\subsection{Groupwise Multi-Task Learning Affinity}
\label{subsec:group_affinity}
Our next goal is to predict affinity among an arbitrary number of tasks in a group. To quantify \textit{groupwise-affinity}, we use \textit{relative MTL gain}, i.e., the relative improvement in prediction performance when several tasks ($t_i, t_j, \cdots $) are trained together using MTL instead of individually with STL, i.e., $\frac{\sum\text{STL loss} - \text{MTL loss}{(t_i,t_j,\cdots)}}{\sum\text{STL loss}}$. 
We explore group relation features that capture the relatedness of multiple tasks within a group alongside the useful task-relation features from our pairwise affinity study.

\pgfplotstableread[col sep=comma]{data/Groupwise_Individual_Usefulness_School_avg_NN.csv}\nnGroupwiseSch
\pgfplotstableread[col sep=comma]{data/Groupwise_Individual_Usefulness_Chemical_avg_NN.csv}\nnGroupwiseChem
\pgfplotstableread[col sep=comma]{data/Groupwise_Individual_Usefulness_Landmine_avg_NN.csv}\nnGroupwiseLM
\pgfplotstableread[col sep=comma]{data/Groupwise_Individual_Usefulness_Parkinsons_avg_NN.csv}\nnGroupwisePK

\begin{figure}[t]
    \begin{subfigure}[b]{0.286\linewidth}
    \begin{tikzpicture}[]
            \begin{axis}
            [
                MediumBarPlot,
                width = \linewidth,
                height = 3.25cm,
                ylabel = Usefulness (\textit{$R^2$}),
                label style = {align = center, font = \footnotesize},
                grid=major,
                yticklabel=\pgfmathprintnumber{\tick}{$\%$},,
                xtick = data,
                ymin = -1,
                xmin = -0.45, xmax = 8.6,
                xticklabels from table={\nnGroupwiseSch}{Feature},
              x tick label style={font = \tiny,rotate = 90, anchor=east},
                legend pos=north east,font = \tiny,
              legend style={at={(0,1)}, anchor=south east}
            ]
            \addplot [TealBars] table [y={Avg_R_SQUARE}, x expr=\coordindex] {\nnGroupwiseSch};
            \end{axis}
        \end{tikzpicture}
        \caption{\textbf{School}}
        \vspace{-0.5pt}
        \label{subfig:groupPredictor_sch}
    \end{subfigure}%
    \begin{subfigure}[b]{0.238\linewidth}
        \begin{tikzpicture}[]
                \begin{axis}
                [
                    MediumBarPlot,
                    width = 1.2\linewidth,
                    height = 3.25cm,
                    label style = {align = center, font = \footnotesize},
                    grid=major,
                    yticklabel=\pgfmathprintnumber{\tick}{$\%$},
                    xtick = data,
                    ymin = -1,
                    xmin = -0.45, xmax = 8.6,
                    xticklabels from table={\nnGroupwiseChem}{Feature},
                  x tick label style={font = \tiny,rotate = 90, anchor=east},
                    legend pos=north east,font = \tiny,
                  legend style={at={(0,1)}, anchor=south east}
                ]
                \addplot [TealBars] table [y={Avg_R_SQUARE}, x expr=\coordindex] {\nnGroupwiseChem};
                \end{axis}
            \end{tikzpicture}
            \caption{\textbf{Chemical}}
            \label{subfig:groupPredictor_chem}
        \end{subfigure}%
        \begin{subfigure}[b]{0.238\linewidth}
        \begin{tikzpicture}[]
                \begin{axis}
                [
                    MediumBarPlot,
                    width = 1.2\linewidth,
                    height = 3.25cm,
                    label style = {align = center, font = \footnotesize},
                    grid=major,
                    yticklabel=\pgfmathprintnumber{\tick}{$\%$},,
                    xtick = data,
                    ymin = -1,
                    xmin = -0.45, xmax = 8.6,
                    xticklabels from table={\nnGroupwiseLM}{Feature},
                  x tick label style={font = \tiny,rotate = 90, anchor=east},
                    legend pos=north east,font = \tiny,
                  legend style={at={(0,1)}, anchor=south east}
                ]
                \addplot [TealBars] table [y={Avg_R_SQUARE}, x expr=\coordindex] {\nnGroupwiseLM};
                \end{axis}
            \end{tikzpicture}
            \caption{\textbf{Landmine}}
            \label{subfig:groupPredictor_lm}
        \end{subfigure}%
        \begin{subfigure}[b]{0.238\linewidth}
        \begin{tikzpicture}[]
                \begin{axis}
                [
                    MediumBarPlot,
                    width = 1.2\linewidth,
                    height = 3.25cm,
                    label style = {align = center, font = \footnotesize},
                    grid=major,
                    yticklabel=\pgfmathprintnumber{\tick}{$\%$},
                    xtick = data,
                    ymin = -1,
                    xmin = -0.45, xmax = 8.6,
                    xticklabels from table={\nnGroupwisePK}{Feature},
                  x tick label style={font = \tiny,rotate = 90, anchor=east},
                    legend pos=north east,font = \tiny,
                  legend style={at={(0,1)}, anchor=south east}
                ]
                \addplot [TealBars] table [y={Avg_R_SQUARE}, x expr=\coordindex] {\nnGroupwisePK};
                \end{axis}
            \end{tikzpicture}
            \caption{\textbf{Parkinson}}
            \label{subfig:groupPredictor_pk}
        \end{subfigure}
\caption{Usefulness of features for predicting relative MTL gain when several tasks ($>2$) are trained together. The average pairwise MTL gain is one of the most effective in predicting relative MTL gain for a group.}
\label{fig:predictor_groupwise}
        \vspace{-2em}
\end{figure}
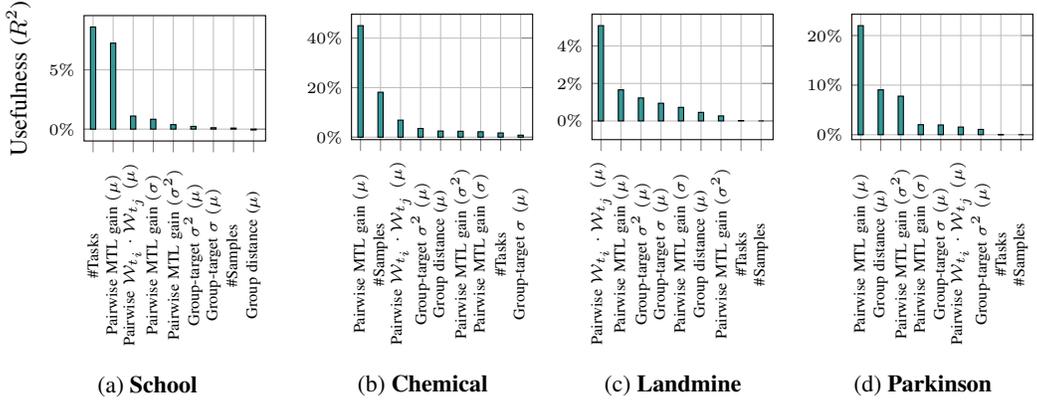

\textbf{Groupwise-Task Features: }
These features describe the relationship between several ($>2$) tasks in a group.
Groupwise-task features that we study include the number of tasks, the average sample size of tasks, the average target variance, and the average target standard deviation of each task. We also consider the  pairwise relative MTL gains for all possible task pairs in a group and use their mean, variance, and standard deviation as groupwise-task features. 
For each pair of tasks in a group, we use the average of the dot products of task-specific parameter vectors ($\mathcal{W}_{t_{i}} \cdot \mathcal{W}_{t_{j}}$) as another groupwise-task feature. 
For distance measures, when studying pairwise-affinity (\cref{subsec:pairwise_affinity}), we considered both distances between the samples of two tasks as well as distances among their combined samples. %
However, we consider only the former for group-relation features since computing distances after combining the samples of several tasks would incur prohibitive computation costs. We also omit the \textit{inter-task affinities} ($\mathcal{Z}_{t_{i}\leftrightarrow t_{j}}$) defined by \citet{fifty2021efficiently} from our groupwise affinity analysis due to their high computational cost and poor predictive performance for predicting pairwise MTL gain.

\textbf{Experiments:} 
We randomly draw a large number of task partitions from the set of all possible partitions with uniform probability for each benchmark;
we choose the number of partitions drawn  based on the number of tasks in each benchmark. 
Specifically, to ensure consistency in the number of groups across the benchmarks, we draw $65$, $200$, $230$, and $150$ partitions for School, Chemical, Landmine, and Parkinson datasets, respectively. 
We train an MTL model for each group with $>2$ tasks, and we
compute the relative MTL gain from the total MTL loss and the previously computed STL losses. To investigate whether the groupwise-task features explain relative MTL gain, we measure the correlation between each feature and relative MTL gain. We perform individual hyperparameter search for each groupwise-task feature to measure each feature's predictive performance ($R^2$).

\textbf{Findings:} \cref{fig:predictor_groupwise} shows the coefficient of determination ($R^2$ value) when predicting relative MTL gain using each groupwise-task feature. We observe that the average pairwise MTL gain for all task pairs in a group is a useful predictor for all benchmarks. 
We also see the dot products of the task-specific parameter vectors ($\mathcal{W}_{t_{i}} \cdot \mathcal{W}_{t_{j}}$) to be useful for three benchmarks out of four.
The Pearson correlation between groupwise-task features and relative MTL gains reveals a consistent pattern closely resembling the measure of usefulness (\cref{appendix:groupaffinity_pearson}).

According to \citet{standley2020tasks}, the accuracy of networks trained on three or more tasks can be predicted by a simple average of the performance of networks trained on pairs of tasks. However, our study demonstrates that a groupwise MTL gain predictor using the average pairwise MTL gain is consistently superior to the simple average in all four benchmark datasets. For example, in the Chemical dataset, the simple average achieves an $R^2$ value of $40.38\%$, while the trained predictor attains an $R^2$ value of $43.35\%$. In the School dataset, the variance in the relative MTL gain values cannot be adequately explained by the simple average of pairwise MTL gains.
In the case of the School, Landmine, and Parkinson benchmarks, the $R^2$ values obtained using a trained predictor are $8.61\%$, $1.66\%$, and $21.99\%$, respectively. This suggests that relying solely on the simple average of pairwise MTL gains is insufficient for predicting higher-order affinities among tasks.

\section{Task Grouping Approach}
\label{sec:approach}
With an understanding of features that are relevant to predicting the success of MTL for a group of tasks, we now focus on the problem of task grouping, i.e., we seek to partition tasks into groups that result in minimum overall loss. One natural approach is to consider all possible subsets of tasks, train MTL models, and then select the partition with the minimum loss.
However, such an exhaustive search approach becomes infeasible as the number of tasks grows. Although \citet{standley2020tasks} suggest heuristic approaches for scalability, e.g., predicting the performance of higher-order networks (with three or more tasks) using the performance
of lower-order networks (with two tasks), an exhaustive search is nonetheless infeasible since even a simple enumeration of possible partitions would take prohibitively long for some benchmarks.
To address this challenge, we propose a novel approach for efficient task grouping. %
We begin by stating our problem formally.

\textbf{Problem Formulation:} Given a set of $n$ tasks $\taskSet = \{t_1,t_2,\dots,t_n\}$, find a partition of the tasks $\taskSet$ into groups $G = \{g_1, g_2, \dots, \}$ that minimizes overall loss $\loss$ when MTL is applied to each group $g \in G$. By partition, we mean that each group $g \in G$ is a subset of tasks $\taskSet_{g} \subseteq \taskSet$, groups are disjoint (i.e., $\taskSet_{g_i} \cap \taskSet_{g_j} = \emptyset$), and their union is the task set $\taskSet$ (i.e., $\bigcup_i \taskSet_{g_i} = \taskSet$).
Formally, we have to solve:
\begin{equation*}
\textstyle
\loss^* = \min_{G \in P(\taskSet)} \sum_{g \in G}\loss_g (\taskSet_g)
\end{equation*}
where $\loss_g(\taskSet_g)$ denotes the total loss for MTL with tasks in $\taskSet_g$ (i.e., sum of loss over tasks in $\taskSet_g$) and $P(\taskSet)$ is the set of all possible partitions of $\taskSet$.

\textbf{Task-Affinity Predictor:} We aim to guide a randomized local search to convergence by reducing the number of MTL trainings with the help of a task-affinity predictor. Specifically, based on a set of groupwise-task features, we design a neural-network predictor of relative MTL gain for a group of tasks (see details on the predictor in \cref{subsec:experimental_setup_results}).

\subsection{Randomized Local Search with Quick Reject}
\label{subsec:quick_reject}
First, we describe our randomized local search approach without the task-affinity predictor, and then we describe how the predictor can speed up the search (\cref{alg:tgp}).

\begin{wrapfigure}{L}{0.5\textwidth}
    \begin{minipage}{0.5\textwidth}
    \vspace*{-.8cm}
\begin{algorithm}[H]
\caption{\textbf{Task Grouping with Predictor}}
\label{alg:tgp}
\textbf{Input}: $\taskSet \gets \{t_1,t_2, \ldots, t_n\}$, 
$\gamma^{retrain}$, $\gamma^{max}$, $\pi^t$

\begin{algorithmic}[1] %
 
\STATE \textbf{TrainPredictor }(\groupSet)\label{alg:line:tgp_train}
\STATE $G \gets \argmin \,\groupSet$; 
$\loss \gets \sum_{g\in G}$ \label{alg:line:minloss_partition}
\FOR{$i = 0, 1, \cdots, \gamma^{max}$ }\label{alg:line:rs_start}
    \STATE $G' \gets$ \textbf{MutateGroups} $(G, \taskSet)$\label{alg:line:group_mutation}
    
    \STATE $\loss_{predicted} \gets$ \textbf{QueryPredictor }$(G')$ \label{alg:line:tgp_query}
    
    \IF{$\loss_{predicted} < \loss$} \label{alg:line:train_or_not}
        \STATE $\textit{train}_{MTL} \gets$ \textbf{True}
    \ELSE
        \STATE $\textit{train}_{MTL} \gets \textbf{RandomBoolean}(\pi^t)$ 
    \ENDIF \label{alg:line:train_or_not_end}
    
    \IF{$\textit{train}_{MTL}$}
        \STATE $\loss' \gets \sum_{g'\in G'}$ \textbf{TrainAndEvaluate}$(\taskSet_{g'})$ \label{alg:line:tgp_train_mtl}
           
        \STATE $\pi \gets \exp\left((\loss- \loss') \cdot K\right)$\label{alg:line:accept_probability}

       \IF{$~\textbf{RandomBoolean}(\pi)$} %
            \STATE  $G \gets G'$, $\loss \gets \loss'$ \label{alg:line:update_g}
        \ENDIF
        \STATE  \groupSet $\,\gets  \groupSet \cup \{(G^\prime,\loss^\prime)\}$
    \ENDIF
 
    \IF{$i \bmod \gamma^{retrain} == 0$}
        \STATE \textbf{UpdatePredictor }(\groupSet)\label{alg:line:update_tgp}
    \ENDIF
\ENDFOR

\STATE \textbf{return} $G^* \gets \argmin \groupSet$\label{alg:line:return}
\end{algorithmic}
\end{algorithm}
    \vspace*{-.8cm}
\end{minipage}
  \end{wrapfigure}

\subsubsection{Randomized Local Search}
We introduce a randomized local search that searches over various partitions of $\taskSet$, trying to minimize overall MTL loss~\loss. 
We initialize our search with the random partitions used for our groupwise MTL affinity study (\textit{Experiments} in \cref{subsec:group_affinity}), which we refer to as our \textit{partition sample}, $\groupSet$.
We have already performed MTL for each group in $\groupSet$ and computed the overall loss for each partition. The partition that attains the lowest loss is selected as the initial partition $G$ with loss $\loss$ (line \ref{alg:line:minloss_partition}).
Then, in each iteration, we generate a new partition $G^\prime$ using a random mutation of the current partition $G$ (line \ref{alg:line:group_mutation}).
In the mutation step, a task $t$ is randomly chosen and moved from its current group $G_{old}$ to a new one $G_{new}^\prime$. 
Based on the cardinality of $G_{old}$, $G_{new}^\prime$ can be either a new group or an existing one in $G$ (details in \cref{appendix:group_mutation}). 
After the mutation, we train MTL models for the modified groups in $G^\prime$ and compute overall loss $\loss^\prime$ (line \ref{alg:line:tgp_train_mtl}).
We \textit{always} accept the new partition $G'$ if its loss is lower than the current one $G$; otherwise, we accept the new partition with a probability that depends on the difference between the loss values and an exogenous parameter $K$ (line \ref{alg:line:accept_probability}).
After a certain number of iterations, we output the best partition $G^*$ found so far (line~\ref{alg:line:return}).

\subsubsection{Randomized Local Search with Affinity Predictor}

The performance bottleneck for our randomized search is evaluating  loss $\loss'$,  which entails MTL training for the mutated groups in every iteration. To alleviate this bottleneck, we extend our randomized search with an affinity predictor that helps to reject inferior partitions quickly. Our extension is based on the idea that a predictor of relative MTL gain, initially trained on a small number of random groups and receiving periodic updates during the search, can guide the randomized search toward better partitions by quickly identifying groups with potentially higher gain values. %

We first train the MTL-gain predictor on groups with more than $2$ tasks from our \textit{partition sample}, $\groupSet$ (line \ref{alg:line:tgp_train}), for which we have already performed MTL, providing ground truth on the MTL-gain values. After each mutation, we first apply the predictor to the new partition  $G'$ and calculate an estimated overall loss $\loss_{predicted}$ from the predicted gains (line \ref{alg:line:tgp_query}). 
If the estimated loss $\loss_{predicted}$ is lower than the current loss $\loss$, then we proceed to MTL training for the mutated groups; otherwise, we proceed only with some probability $\pi^t$ (lines \ref{alg:line:train_or_not}-\ref{alg:line:train_or_not_end}).
By quickly rejecting unpromising mutations $G'$ with probability $1-\pi^t$, we significantly improve the performance of randomized search.
Further, to improve the MTL-gain predictor over time, we retrain the predictor after every $\gamma^{retrain}$ iterations, using all the groups of all the partitions $\groupSet$ for which we have performed MTL during search (line \ref{alg:line:update_tgp}).

\textbf{Computational Advantage:}
The cost of the initial MTL training and the occasional task-affinity predictor training is negligible compared to performing multi-task learning in each iteration of the search. For example, in the School benchmark with 2 groups of 13 and 15 tasks, the total MTL training time is more than 3.2 minutes, while the total time for training and querying the task-affinity predictor is only 0.37 minutes (we report results for the other benchmarks in \cref{appendix:queryTime}).

\section{Numerical Evaluation of Task Grouping Approach}
\label{sec:result}
Here, we demonstrate the effectiveness of our proposed search approach on the benchmark datasets.

\subsection{Experimental Setup for Task Grouping}
\label{subsec:experimental_setup_results}

We first perform an individual hyper-parameter and feature-selection search for each benchmark to determine the architecture and input features of the task-affinity predictor. 
We build the affinity predictor using the best-performing architecture from the search (feed-forward neural networks with multiple hidden layers and non-linear activation functions; more details in \cref{appendix:affinity_pred_architecture}). The search and initial training of the task-affinity predictor are performed using the groups from \textit{partition sample}, $\groupSet$ (\textit{Experiments} in \cref{subsec:group_affinity}). We also initialize our task-grouping search with partitions from $\groupSet$, where we start each execution from a different initial partition $G$, such as the best, second best, or third best partition within $\groupSet$.
We set $\gamma^{retrain}$ to 5, indicating that the predictor is updated after every 5 iterations of our search algorithm. Early in the search, we proceed to MTL training for a mutation $G'$ with probability $\pi^t = 0.1$ (i.e., 10\%) when $\loss_{predicted}$ is higher than the current $\loss$, and we decrease $\pi^t$ over time as the predictor improves its ability to reject mutations.
We measure the computational cost of our approach in terms of the total number of MTL models trained. Since we run our experiments on a multi-user computer cluster, it is difficult to reliably measure actual compute time (due to resource scheduling and priorities); as a result, we consider the number of MTL models trained, which serves as a sufficiently accurate estimate of computational cost.

\subsection{Baseline Approaches}
\label{sec:baseline_approaches}
We compare to several baseline methods to demonstrate the effectiveness of our proposed approach.

\subsubsection{Exhaustive Search}
Prior work on task grouping mostly focused on exhaustive search~\citep{standley2020tasks,fifty2021efficiently} for datasets with fewer tasks. An exhaustive search would be infeasible for our benchmark datasets since it would take years to enumerate all the possible partitions.
For example, the Parkinson dataset has 42 tasks, so the number of all possible partitions is \num[round-precision=3,round-mode=figures]{35742549198872617291353508656626642567}. 
Even if we could somehow estimate the loss for each partition in a nanosecond, an exhaustive search would take over $10^{21}$ years (by which time the Earth is expected to have been consumed by the Sun).
Since a simple exhaustive search is infeasible, to enable comparison, we substitute it with a simple random search that tries random partitions one-by-one, until it has used at least the same amount of computation as our approach (number of MTL models trained), and then it outputs the best partition found so far.

\subsubsection{Clustering Algorithms}
Motivated by prior work~\cite{he2019efficient}, we also apply classic clustering algorithms, requiring that tasks within the same cluster have high affinity.
We explore multiple clustering approaches for obtaining different task partitions and compare them with our proposed approach.

\begin{table}[t!]
    \centering
    \caption{Comparison of Different STL and MTL Models and Grouping Approaches on Benchmark Datasets\newline\emph{$^*$Note that since exhaustive search is very far from being computationally feasible, we substitute it with a random search. Our approach consistently outperforms all baselines across all benchmark datasets.}}
    \label{tab:comparison_with_SOTA_MTL}
    \vspace{-0.025em}
    \setlength{\tabcolsep}{2.5pt}
    \renewcommand{\arraystretch}{1.15}
    \resizebox{\linewidth}{!}{%
    \begin{tabular}{*{11}{|c}|}
    \hline
     &  & &  & \textbf{Pairwise MTL} & \textbf{Pairwise MTL} & \textbf{Simple} &\multicolumn{2}{c|}{\textbf{Clustering Algorithms}}&\textbf{Exhaustive$^*$} & {\textbf{Our Approach}}\\ \cline{8-9} 
     
    \textbf{Dataset} & \textbf{Tasks} & \textbf{Evaluation} & \textbf{STL} & (all pairs) &  (optimal) & \textbf{MTL} &\textbf{Hierarchical}& \textbf{$k$-Means}& (Random & (Random Search\\
    
     & \textbf{$n$} & \textbf{Metric} & $|g_i|=1$  & $|g_i|=2$  & $|g_i|=2$ & $|g_i|=n$ &(MTL Affinities)&($\mathcal{W}_{t_i} \cdot\mathcal{W}_{t_j}$)& Partitionings)& w. Quick Reject)\\ \hline \hline
    
    \textbf{School} &139 & $\sum$\textit{MSE} $\downarrow$ & 86.49 & 92.58 & 75.71 & 116.29 $\pm$ 1.35 
    &102.96 $\pm$ 0.58
    & 98.43 $\pm$ 0.33
    & 91.05 $\pm$ 0.1
    & \textbf{78.7 $\pm$ 0.7}\\ \cline{3-11}
     \hline
    
    \textbf{Chemical} &35 &$\sum$\textit{log-loss} $\downarrow$ & 18.66  & 17.75 & 13.60 & 18.44 $\pm$ 0.11 
    &17.27 $\pm$ 0.14
    & 17.49 $\pm$ 0.0
    & 16.85 $\pm$ 0.0
    & \textbf{15.09 $\pm$ 0.08}\\ \cline{3-11}
    
     & & error $\downarrow$ & 0.33  & 0.31 & 0.27 & 0.31 $\pm$ 0.002 
     & 0.27 $\pm$ 0.02
     & 0.28 $\pm$ 0.0
     & 0.27 $\pm$ 0.0
     & \textbf{0.12 $\pm$ 0.0}
     \\ \hline
    
    \textbf{Landmine} & 29 &$\sum$\textit{log-loss} $\downarrow$ & 5.56 & 5.63 & 5.08 
    &  5.63 $\pm$  
    & 5.3 $\pm$ 0.04 
    &  5.35 $\pm$ 0.01
    &5.17 $\pm$ 0.0
    &\textbf{4.9 $\pm$ 0.04}\\ \cline{3-11}
    
     & & \textit{AUC} $\uparrow$ & 73.86\% & 76.4\% & 78.56\%& 78.27\% $\pm$ 2\% 
     & 84.73\% $\pm$ 0.02\%
     & 87.92\% $\pm$ 0.0\%
     &79.09\% $\pm$ 0.0\%
     &\textbf{90.19\% $\pm$ 0.01\%} \\\hline

    \textbf{Parkinson} &42 & $\sum$\textit{MSE} $\downarrow$ & 281.50 & 269.64 & 240.91 & 276.13 $\pm$ 0.78  
    & 259.08 $\pm$ 1.95
    & 268.41 $\pm$ 0.45
    & 254.85 $\pm$ 0.0
    & \textbf{233.87 $\pm$ 0.52}\\\hline
    \end{tabular}
    }
    \vspace{-1.5em}
\end{table}

\noindent \textbf{Hierarchical Clustering:}
We use \citet{shiri2022highly}'s hierarchical clustering approach with \emph{a priori} information describing the task relations in a hierarchical structure. We describe task relations using pairwise MTL gains and task-specific network parameters (dot products, $\mathcal{W}_{t_i}\cdot\mathcal{W}_{t_j}$) from the simple MTL model. We explore various combinations of hyper-parameters for the clustering algorithm, resulting in $137, 113, 112,$ and $152$ distinct partitionings for School, Chemical, Landmine, and Parkinson datasets, respectively, with different numbers of clusters. For each cluster, we train and evaluate MTL models, and this training-evaluation process is repeated multiple times.

\noindent  \textbf{$k$-Means Clustering:}
For $k$-means clustering, we represent each task as a vector of task-specific network parameters from the simple MTL model (\citet{shiri2022highly}.
Using the \textit{elbow} method, we identify the optimal range of the number of clusters, $k$, and obtain $24$, $18$, $14$, and $19$ distinct partitionings with varying numbers of clusters for the School, Chemical, Landmine, and Parkinson datasets, respectively. Each partitioning is evaluated in a similar manner to hierarchical clustering.

\subsubsection{Other Approaches}

We also compare our proposed approach with simple but interesting baselines: \textbf{Single Task Learning (STL)} ($|g_i| = 1$), \textbf{Pairwise MTL} ($|g_i| = 2$), and \textbf{Simple MTL} ($|g_i| = n$). For STL, each task is trained independently. For Pairwise MTL, we train separate MTL models for each pair of tasks and measure each task's loss by considering either the mean loss over all its pairings or its optimal pairing. For Simple MTL, we train all tasks in a group using a single hard parameter-sharing MTL model.

\subsection{Comparison}

In \cref{tab:comparison_with_SOTA_MTL}, we report the performance of our proposed approach and compare it against the baselines. The reported results for each approach have been gathered through multiple repetitions of MTL training and evaluation process. The average loss over all tasks after STL ($|g_i| = 1$) and pairwise MTL ($|g_i| = 2$) demonstrates the usefulness of MTL. We see significant improvements even when considering all possible pairings instead of only the ones that had a positive transfer of information during training. The results from MTL with all tasks trained together ($|g_i| = n$) show improvement over STL, but they also demonstrate that negative transfer of information can reduce performance when we put dissimilar tasks in the same group (compared to pairwise MTL). For classical task-clustering algorithms, we report average losses over all the partitions in \cref{tab:comparison_with_SOTA_MTL}, but the losses of the best partitions (76.86 (School \textit{MSE}), 14.87 (Chemical \textit{log-loss}), 4.78 (Landmine \textit{log-loss}), and 232.28 (Parkinson \textit{MSE})) are \emph{still significantly outperformed by our approach}.
Our approach also \emph{significantly outperforms} the simple random search (which serves as a substitute for the computationally infeasible exhaustive search). 
Our approach also exhibits similar superior performance for MTL models based on SVM and XGBoost (\cref{appendix:extra_results}).

\section{Conclusion}
\label{sec:concl}

To improve the application of multi-task machine learning, %
we investigate the fundamental questions of \emph{how to predict if a group of tasks would benefit from MTL} and \emph{how to efficiently find groups that maximize the benefits of MTL}. 
By studying four different benchmark datasets, we find that features such as STL curve gradients and dataset size are useful for predicting relative MTL gain when two tasks are trained together; we also find that average pairwise MTL gain and average dot product of task-specific parameters are good predictors of group affinity.
To automate partitioning tasks for MTL, we introduce a scalable randomized search algorithm, which significantly outperforms existing baseline approaches.

\clearpage

\addcontentsline{toc}{section}{References}
\bibliographystyle{named}
\bibliography{main}

\clearpage
\appendix

\section{Dataset Description}
\label{appendix:dataset_desc}
We employ four widely used benchmark datasets in our study~\citep{zhang2021survey}. We summarize the characteristics of these datasets in \cref{tab:datadesc}. 

\begin{table}[h]
\setlength{\tabcolsep}{5pt}
\caption{Benchmark Datasets}
\def\arraystretch{1.1}
\centering
\begin{tabular}{|l|r|r|r|r|c|r|r|}
\hline %
\multirow{2}{*}{\textbf{Dataset}}  &
\multirow{2}{*}{\textbf{Task Type}}  &
\multirow{2}{*}{\textbf{Loss Metric}}  &
\multirow{2}{*}{\#\textbf{Tasks}}  &
\multirow{2}{*}{\#\textbf{Features}}  &
\multicolumn{3}{c|}{\textbf{Task Sample Size}} \\ \cline{6-8}
&    &  & &    & Avg.     & Min.     & Max.     \\ \hline %
\hline
School   & Regression  & MSE & 139 & 22 & 110& 22& 251\\ \hline %
Chemical & Classification  & \textit{log-loss, error rate} & 35 & 180 & 435 & 22 & 2,368\\ \hline %
Landmine & Classification  & \textit{log-loss}, AUC& 29 & 9 & 511 & 445 & 690 \\ \hline
Parkinson & Regression  & MSE& 42 & 16 & 139 & 101 & 168 \\ \hline
\end{tabular}
\label{tab:datadesc}
\end{table}

\paragraph{Loss Metrics}
Since the four benchmarks span diverse prediction problems (e.g., regression vs.\ classification), we use a diverse set of metrics to report performance. For all of our experiments, we minimize distinct loss metrics for different task types, i.e., \textit{mean squared error (MSE)} for regression tasks and \textit{binary cross entropy} (\textit{log-loss}) for classification tasks. In addition, for comparison with prior work \citep{zhang2021survey}, we also report two additional metrics for the Chemical and Landmine benchmarks: \emph{error rate} and \emph{area under the curve (AUC)}.

\paragraph{Sample Distribution for Multi-Task Learning}
In multi-task learning, equal sample distribution over the tasks facilitates effective parameter sharing, prevents biased gradient updates, and promotes better generalization. Further, it also mitigates overfitting tasks with fewer samples. Since tasks within the same benchmark have varying numbers of training samples (see \emph{Task Sample Size} in \cref{tab:datadesc}), we apply data augmentation for each benchmark to help address the data imbalance. Specifically, we repeat the training samples to match the maximum sample size (\cref{tab:datadesc}) for all MTL scenarios to maintain the same number of examples across all tasks.

Furthermore, we keep the same number of positive and negative samples for the Chemical dataset, similar to prior work \citep{jacob2008clustered}. When implementing Regularized Multi-task Learning (RMTL) with a support vector machine, we opt for a smaller sample size (maximum $500$ with an equal number of positive and negative samples from each task) for the Chemical dataset because of the high computational cost associated with the optimization process of RMTL with SVM, particularly when dealing with larger datasets.

\section{Feature Description}
\label{appendix:feature_desc}

\subsection{Pairwise Features}
\label{appendix:pairwise_feature_desc}

This subsection describes the features that we use in our task-affinity study with pairs of tasks, $t_i$ and $t_j$. For reference, we present basic notation in \cref{tab:notation}, which we will use throughout the description.

\begin{table}[]
\def\arraystretch{1.15}
\centering
\caption{Notation for Features}
\label{tab:features}
\begin{tabular}{|c|c|}
\hline
\textbf{Symbol}  & \textbf{Description} \\
\hline\hline
$\sigma$ & Standard Deviation \\ \hline
$\sigma^2$ & Variance \\ \hline
$\mu$ & Average \\ \hline
$D_t$ & Dataset (i.e., sample) size of task $t$ \\ \hline
$d_E$ & Euclidean distance \\ \hline
$d_M$ & Manhattan distance \\ \hline
$d_H$ & Hamming distance \\ \hline
$\mathcal{W}_t$ & Network parameters of a task-specific layer \\ \hline

nS & \textbf{Normalized} by total \textbf{sum} \\ \hline
nP & \textbf{Normalized} by total \textbf{product} \\ \hline

$(t_i \leftrightarrow t_j)$ & \textbf{Between tasks} $t_i$ and $t_j$\\ \hline
$(t_i + t_j)$ & \textbf{Combined samples} of tasks $t_i$ and $t_j$\\ \hline
scaled & After \textbf{feature-scaling} \\ \hline

\end{tabular}
\label{tab:notation}
\end{table}

We begin by listing simple features that pertain to the datasets of a pair of tasks $t_i$ and $t_j$:
\begin{itemize}
    \item \textbf{$|D_{t_i}-D_{t_j}|$ nS}: difference between the dataset sizes of $t_i$ and $t_j$, normalized by their sum, i.e.,  $\frac{|D_{t_i} -D_{t_j}|}{D_{t_i}+D_{t_j}}$.

    \item \textbf{Target} $\sigma$ ($\mu$): average of the standard deviations $\sigma_{t_i}$ and $\sigma_{t_j}$ of the target attributes of $t_i$ and $t_j$.

    \item \textbf{$|\sigma_{t_i}-\sigma_{t_j}|$ nS}: difference in the $\sigma$ of the target attributes of $t_i$ and $t_j$, normalized by their sum sum, i.e., $\frac{|\sigma_{t_i}-\sigma_{t_j}|}{\sigma_{t_i}+\sigma_{t_j}}$.
    
    \item \textbf{$\sigma_{(t_i + t_j)}$ nS}: $\sigma$ of the target attribute after combining (i.e., taking the union of) the samples of tasks $t_i$ and $t_j$, normalized by the sum of $\sigma_{t_i}$ and $\sigma_{t_j}$, i.e., $\frac{\sigma_{(t_i+t_j)}}{\sigma_{t_i}+\sigma_{t_j}}$.
    
    \item \textbf{$\sigma_{(t_i + t_j)}$ nP}: $\sigma$ of the target attribute  after combining the samples of tasks $t_i$ and $t_j$, squared and then normalized by the product of $\sigma_{t_i}$ and $\sigma_{t_j}$, i.e., $\frac{(\sigma_{(t_i+t_j)})^2}{\sigma_{t_i} \cdot \sigma_{t_j}}$.
    
    \item \textbf{Target} $\sigma^2$ ($\mu$): average of $\sigma_{t_i}^2$ and $\sigma_{t_j}^2$ of the target attributes of $t_i$ and $t_j$.
    
    \item \textbf{$|\sigma^2_{t_i}-\sigma^2_{t_j}|$ nS}: difference in $\sigma^2$ of the target attributes, normalized by their sum, i.e., $\frac{|\sigma^2_{t_i}-\sigma^2_{t_j}|}{\sigma^2_{t_i}+\sigma^2_{t_j}}$.
    
    \item \textbf{$\sigma^2_{(t_i + t_j)}$ nS}: $\sigma^2$ of the target attribute after combining the samples of tasks $t_i$ and $t_j$, normalized by the sum of $\sigma_{t_i}^2$ and $\sigma_{t_j}^2$, i.e., $\frac{\sigma^2_{(t_i+t_j)}}{\sigma^2_{t_i} + \sigma^2_{t_j}}$.
    
    \item \textbf{$\sigma^2_{(t_i + t_j)}$ nP}: $\sigma^2$ of the target attribute after combining the samples of tasks $t_i$ and $t_j$, squared and then normalized by the product of $\sigma_{t_i}^2$ and $\sigma_{t_j}^2$, i.e., $\frac{({\sigma^2_{(t_i+t_j)}})^2}{\sigma^2_{t_i} \cdot \sigma^2_{t_j}}$.
\end{itemize}

\paragraph{STL-based Features}
Next, we list features that are based on single-task learning performance:
\begin{itemize}
    \item \textit{\textbf{Curve grad}}$_{t_i} (x\%)$: gradient of the loss curve for task $t_i$ from its single-task training, computed at $x\%$ completion of the training process. The gradient is computed as the relative change in the test loss at $x\%$ of the training compared to the test loss at the end of training.

    \item \textbf{Curve grad} $ (x\%)$: normalized difference between the gradients of the STL loss curves of tasks $t_i$ and $t_j$, computed at $x\%$ completion of their training.

    \item \textbf{Fitted param a} $t_i$: parameter $a$ of a logarithmic curve fitted to the STL loss curve of task $t_i$ (see \cref{subsec:pairwise_affinity}).

    \item \textbf{Fitted param b} $t_i$: parameter $b$ of a logarithmic curve fitted to the STL loss curve of task $t_i$ (see \cref{subsec:pairwise_affinity}).

    \item \textbf{Param $|a_{t_i} - a_{t_j}|$}: difference in the parameters $a$ of the logarithmic curves fitted to the STL loss curves of tasks $t_i$ and $t_j$, normalized by the sum of these parameter values.

    \item \textbf{Param $|b_{t_i} - b_{t_j}|$}: difference in the parameters $b$ of the logarithmic curves fitted to the STL loss curves of tasks $t_i$ and $t_j$, normalized by the sum of these parameter values.

    \item \textbf{\textit{Inter-Task Affinity} ($\mathcal{Z}_{t_i \leftrightarrow t_j}$)}: A measure of how task $t_i$'s gradient update affects task $t_j$'s loss in the simple MTL model. The formulation is based on prior work by \citet{fifty2021efficiently}.

    \item \textbf{($\mathcal{W}_{t_i} \cdot \mathcal{W}_{t_j}$) }: Dot product of vectors $\mathcal{W}_{t_i}$ and $\mathcal{W}_{t_j}$, containing task-specific network parameters from the simple MTL model. Specifically, for any task~$t$, $\mathcal{W}_{t}$ consists of the weights and biases of the task-specific output layer from the simple MTL model, which includes all tasks.

\end{itemize}

Note that we use these STL-based features only for MTL models trained with neural networks. Since non-linear SVMs are typically not trained using gradient-based methods, we could not gather the learning curve-based features for SVM. Although XGBoost trees follow a gradient-based optimization with decision trees to minimize the loss, calculating the loss at various stages of the training presents technical challenges with the widely used implementation of XGBoost. 
As a result, we disregard these parameters for SVM and XGBoost MTL. 
Similarly, the weight matrices in non-linear SVMs are not directly interpretable. Unlike linear SVMs, where the weight matrix corresponds to the separating hyperplane, non-linear SVMs employ kernel functions to map the data into a higher-dimensional feature space. As a result, obtaining the weight matrices for non-linear SVMs is not as straightforward as separating hyperplanes in the original feature space for linear SVMs.

\paragraph{Distance Features}
Finally, we list features that are computed by measuring the average distance within the sample of one task, between the samples of two tasks, and within the combination (i.e., union) of the samples of two tasks.

\begin{itemize} 
    \item $d_{E_{t_i}}$: average Euclidean distance within the sample of task $t_i$ (i.e., average of the Euclidean distances between all pairs of datum  in the dataset of task $t_i$).
    
    \item $d_{M_{t_i}}$: average Manhattan distance within the sample of task $t_i$.
    
    \item $d_{H_{t_i}}$: average Hamming distance within the sample of task $t_i$.
    
    \item $d_{E_{(t_i \leftrightarrow t_j)}}$: average Euclidean distances between the samples of tasks $t_i$ and $t_j$ (i.e., average of the Euclidean distances between pairs of datum from the samples of tasks $t_i$ and $t_j$).
    
    \item $d_{M_{(t_i \leftrightarrow t_j)}}$: average Manhattan distances between the samples of tasks $t_i$ and $t_j$.
    
    \item $d_{H_{(t_i \leftrightarrow t_j)}}$: average Hamming distances between the samples of tasks $t_i$ and $t_j$.
    
    \item \textbf{$d_{E_{(t_i \leftrightarrow t_j)}}$ nS}: average Euclidean distance between tasks $t_i$ and $t_j$, normalized by the sum of $d_{E_{t_i}}$ and  $d_{E_{t_j}}$, i.e, $\frac{d_{E_{(t_i \leftrightarrow t_j)}}}{d_{E_{t_j}} + d_{E_{t_j}}}$.
    
    \item \textbf{$d_{E_{(t_i \leftrightarrow t_j)}}$ nP}: square of the average Euclidean distance between tasks $t_i$ and $t_j$, normalized by the product of $d_{E_{t_i}}$ and  $d_{E_{t_j}}$, i.e., $\frac{d_{E_{(t_i \leftrightarrow t_j)}}^2}{d_{E_{t_i}} \cdot d_{E_{t_j}}}$. 
    
    \item \textbf{$d_{M_{(t_i \leftrightarrow t_j)}}$ nS}: average Manhattan distance between tasks $t_i$ and $t_j$, normalized by the sum of $d_{M_{t_i}}$ and  $d_{M_{t_j}}$, i.e., $\frac{d_{M_{(t_i \leftrightarrow t_j)}}}{d_{M_{t_i}} + d_{M_{t_j}}}$.
    
    \item \textbf{$d_{M_{(t_i \leftrightarrow t_j)}}$ nP}: square of the average Manhattan distance between tasks $t_i$ and $t_j$,  normalized by the product of $d_{M_{t_i}}$ and  $d_{M_{t_j}}$, i.e., $\frac{d_{M_{(t_i \leftrightarrow t_j)}}^2}{d_{M_{t_i}} \cdot d_{M_{t_j}}}$.
    
    \item \textbf{$d_{H_{(t_i \leftrightarrow t_j)}}$ nS}: average Hamming distance between tasks $t_i$ and $t_j$, normalized by the sum of $d_{H_{t_i}}$ and  $d_{H_{t_j}}$, i.e., $\frac{d_{H_{(t_i\leftrightarrow t_j)}}}{d_{H_{t_i}} + d_{H_{t_j}}}$.
    
    \item \textbf{$d_{H_{(t_i \leftrightarrow t_j)}}$ nP}: square of the average Hamming distance between tasks $t_i$ and $t_j$, $d_{H_{(t_i,t_j)}}$ normalized by the product of $d_{H_{t_i}}$ and  $d_{H_{t_j}}$, i.e., $\frac{d_{H_{(t_i \leftrightarrow t_j)}}^2}{d_{H_{t_i}} \cdot d_{H_{t_j}}}$.
    
    \item \textbf{$d_{E_{(t_i + t_j)}}$ nS}: average Euclidean distance measured after combining the samples of tasks $t_i$ and $t_j$ (i.e., average of the Euclidean distances between all pairs of datum in the union of the datasets of tasks $t_i$ and $t_j$), normalized by the sum of the individual distances $d_{E_{t_i}}$ and  $d_{E_{t_j}}$.
    
    \item \textbf{$d_{E_{(t_i + t_j)}}$ nP}: average Euclidean distance measured after combining the samples of tasks $t_i$ and $t_j$, squared and normalized by the product of the individual distances $d_{E_{t_i}}$ and  $d_{E_{t_j}}$.
    
    \item \textbf{$d_{M_{(t_i + t_j)}}$ nS}: average Manhattan distance measured after combining the samples of tasks $t_i$ and $t_j$, normalized by the sum of the individual distances $d_{M_{t_i}}$ and  $d_{M_{t_j}}$.
    
    \item \textbf{$d_{M_{(t_i + t_j)}}$ nP}: average Manhattan distance measured after combining the samples of tasks $t_i$ and $t_j$, squared and normalized by the product of the individual distances $d_{M_{t_i}}$ and  $d_{M_{t_j}}$.
    
     \item \textbf{$d_{H_{(t_i + t_j)}}$ nS}: average Hamming distance measured after combining the samples of tasks $t_i$ and $t_j$, normalized by the sum of the individual distances $d_{H_{t_i}}$ and  $d_{H_{t_j}}$.
     
    \item \textbf{$d_{H_{(t_i + t_j)}}$ nP}: average Hamming distance measured after combining the samples of $t_i$ and $t_j$, squared and normalized by the product of the individual distances $d_{H_{t_i}}$ and  $d_{H_{t_j}}$.

\end{itemize}
For School, Landmine, and Parkinson data, we also consider  variations of these distance features after scaling the data (so that all data features have zero mean and unit variance), and we denote these features with \emph{scaled} in the features' short names.

\subsection{Groupwise Features}
\label{appendix:groupwise_feature_desc}
In this subsection, we present the features that we employ in our task-affinity study for multiple (i.e., more than two) tasks  trained within a group. In addition to the task-relation features derived from our pairwise affinity study, we introduce various group-relation features that capture the relatedness among tasks within a group. 

The following list provides a detailed description of our groupwise task features.

\begin{itemize}
    \item \textbf{\#Tasks:} number of tasks to train together in a group.
    \item \textbf{\#Sample:} average of the sample sizes of each task in a group.
    \item \textbf{Group-target $\sigma^2 (\mu)$:} average variance ($\sigma^2$) of the target attributes of all tasks in a group.
    \item \textbf{Group-target $\sigma (\mu)$:} average standard deviation ($\sigma$) of the target attributes of all tasks in a group.
    \item \textbf{Group Distance ($\mu$):} average distance between all possible pairs of tasks in a group (for School, Landmine and Parkinson dataset, we consider Euclidean distance; and for Chemical benchmark, we consider the Hamming distance between each task pairs).
    
    \item \textbf{Pairwise MTL gain ($\mu$):} average of the pairwise relative MTL gains for all possible task pairs in a group.
    \item \textbf{Pairwise MTL gain ($\sigma^2$):}  variance ($\sigma^2$) of the pairwise relative MTL gains for all possible task pairs in a group.
    \item \textbf{Pairwise MTL gain ($\sigma$):} standard deviation ($\sigma$) of the pairwise relative MTL gains for all possible task pairs in a group.
    \item \textbf{Pairwise $\mathcal{W}_{t_{i}} \cdot \mathcal{W}_{t_{j}} (\mu)$:}  average of the dot products of task-specific parameter vectors ($\mathcal{W}_{t_{i}} \cdot \mathcal{W}_{t_{j}}$) for each pair of tasks in a group .
    
\end{itemize}

\section{MTL Architectures}
\label{appendix:mtl_architectures}
This section describes the general multi-task learning (MTL) architectures that we use for all experiments in the paper. 

\subsection{Neural Network}
We employ feed-forward neural networks for the MTL architectures for regression (School and Parkinson) and classification tasks (Chemical and Landmine). Each neural network architecture consists of a few shared hidden layers across tasks while maintaining task-specific layers.
Input data for each task is fed through the input layer, followed by shared hidden layers capturing common features and task-specific hidden layers capturing task-specific patterns. Each task has a separate output layer of its own. The shared layers promote joint learning during training, while task-specific layers specialize in each task. The network optimizes parameters using a multi-task loss function, improving performance across all tasks. In the output layers of other MTL models, we use \textit{linear} activation and \textit{softmax} activation for regression and classification tasks, respectively. We train the MTL models to minimize \textit{mean squared error (MSE)} for regression tasks and binary cross-entropy \textit{(log-loss) }for classification, and models for both types of tasks are optimized using the Adam optimizer~\cite{kingma2014adam}.

For our final MTL model implementation, we first perform a neural architecture search (NAS) on a randomly chosen subset of tasks for each benchmark. The starting architecture consists of a task-specific pre-processing module (before shared layers), some shared hidden layers, and a task-specific post-processing module (after shared layers). We explore variations in the number of different types of layers, the number of neurons in each layer, the learning rate, and the activation function for the hidden layers using a standard random localized search approach \citep{bergstra2012random,ayman2022neural}. For computational efficiency, we run the search until a  fixed number of iterations and choose the best-performing architectures until that point. Finally, we use these best-performing architectures for all our MTL-based experiments.

\subsection{Regularized Support Vector Machine}

For our multi-task learning models using support vector machines (SVM), we implement Regularized Multi-Task Learning (RMTL) \citep{evgeniou2004regularized}.
Regularized MTL with SVM involves jointly learning multiple tasks using SVMs while incorporating regularization. The optimization problem minimizes task-specific loss functions and optimizes regularization terms to learn shared patterns across tasks. Formulation and implementation details vary based on problems, chosen regularization strategy, and optimization algorithm. 

We utilize the SVM library provided by scikit-learn for implementing Regularized MTL with SVM. We employ a non-linear kernel function called the \textit{radial basis function (RBF)} to compute the kernel matrix. Our approach uses two positive regularization parameters: $\lambda_1$, controlling task-specific weights, and $\lambda_2$, governing shared weights. Additionally, we select two more parameters, $C$ and $\mu$, to handle training error and task similarity, respectively. Since the optimal choice of regularization parameters depends on the dataset and desired model behavior, we perform individual hyper-parameter searches for $\lambda_1$ and $\lambda_2$ in each benchmark. Furthermore, we tune additional hyper-parameters, such as maximum iterations, to put a limit on iterations within the solver and the 
sigma parameter ($\sigma$) to control the level of non-linearity in the \textit{RBF} kernel. Finally, we determine the hyper-parameters for our RMTL with SVM models based on the results of the search.

\subsection{Extreme Gradient Boosting Trees}

Extreme Gradient Boosting (XGBoost) is a gradient boosting framework that can be extended for multi-task learning (MTL). It combines decision trees as weak learners with task-specific loss functions to jointly optimize multiple related tasks. XGBoost enhances the model's predictive performance by leveraging shared information and relationships across tasks. 

For our MTL implementation with XGBoost, we use the XGBoost library \citep{Chen:2016:XST:2939672.2939785}. For each task group, we first convert the training samples from each task into LibSVM format \citep{chang2011libsvm}, which provides a straightforward representation of datasets, and supports both sparse and dense data. It also optimizes memory and computation and is widely supported by machine learning libraries. We perform a random search over various hyper-parameters of the XGBoost trees (number of trees, learning rate, regularization, etc.) and determine the final structure of the model. The MTL models use different error metrics (e.g., squared error, log loss) as their objectives based on different tasks. 

\emph{Please note that the specific MTL architectures are not a crucial contribution to our study. While MTL architectures have been widely studied in various contexts, their inclusion in our study is peripheral and not the primary highlight of our research. Instead, our study primarily emphasizes learning the relationship between tasks and leveraging the knowledge for an efficient task-grouping strategy.}

\section{Neural Architecture and Feature Search for Affinity Predictor}
\label{appendix:NAS}
We perform hyper-parameter and feature-selection searches multiple times throughout our study. This section gives a high-level overview of the hyper-parameter optimization methodology, mostly focusing on the neural architecture and feature-selection search performed for our study of Task Affinity (\cref{sec:affinity}).

\begin{wrapfigure}{L}{0.5\textwidth}
    \begin{minipage}{0.5\textwidth}
    \vspace*{-.5cm}
\begin{algorithm}[H]
\caption{\textbf{Random Local Search} ($\arch$)}
\label{algo:randomNeighbour}
\textbf{Input}: 
$\Omega  \leftarrow$ \{\hyperParam\}, Search Space\\
$\probabilityDistribution_{\hyperParam} \gets$ \{f(\hyperParam):f(\hyperParam)$\geq0$\text{, }$\forall_{\hyperParam}\in\Omega\}$ \\
$\beta_\alpha, \beta_\neuron \gets \text{change percentage of }\alpha \text{, } \neuron$

\begin{algorithmic}[1] %
 \STATE $hp = $ \textbf{RandomChoice}($\Omega, 1, p=\probabilityDistribution_{\hyperParam})$
 
 \IF{$hp == \featureNN$}
    \STATE $\featureNN_{curr} \gets \arch[\featureNN]$
     \STATE $\featureNN^\prime \gets $ \textbf{changeFeatures} ($\featureNN_{curr}$)\;
     \STATE $\arch^\prime[\featureNN] \gets \featureNN^\prime$

    \ELSIF{$hp == \layerNN$}
     \STATE $\arch^\prime \gets $ \textbf{changeLayers} ($\arch$)\;

 \ELSIF{$hp == \neuron$}
     \STATE $h_{layer} := $ \textbf{RandomChoice} ($[0, |\arch [\layerNN]|]$)
    
    \STATE $\neuron_{curr} \gets \arch [\neuron][h_{layer}]$
    
     \STATE  $\neuron^\prime \gets \neuron_{curr} + \lceil
            \neuron_{curr} \cdot \beta_{\neuron} \rceil$
        \STATE  $\arch^\prime[\neuron][h_{layer}] \gets \neuron^\prime$

  \ELSE
    \STATE $\arch^\prime[\alpha] \gets \arch[\alpha] +
           \arch[\alpha]  \cdot \beta_{\alpha}$ 
         
  \ENDIF
\STATE \textbf{return} $\arch^\prime$ 
  \end{algorithmic}
\end{algorithm}
    \vspace*{-0.5cm}
\end{minipage}
  \end{wrapfigure}
We implement feed-forward neural networks for predicting relative MTL gain and conduct hyper-parameter searches for the predictor architectures. 
First, to measure the usefulness of each individual feature (pairwise-task and groupwise-task), we try to predict the relative MTL gain using a model that uses only that feature. Before finalizing the predictor hyper-parameters, we run a hyper-parameter search to find the architecture of the predictor. The hyper-parameter searches aim to maximize prediction performance, measured by $R^2$ value while controlling the model complexity. Model complexity is defined in terms of the model's total number of trainable parameters.  Once the individual usefulnesses of all features are measured, we build a predictor that takes an optimal set of input features and the optimal hyper-parameters for the predictor network. We perform another hyper-parameter and feature-selection search to optimize the architecture and the input predictor variables to the network.

\textbf{Search Space : }
Depending on our objective of the search (architecture's hyper-parameters, input features, or both), our search space $\Omega$ consists of multiple hyper-parameters ($\Omega = \{\hyperParam\}$) for the architecture. $\Omega$ includes the number of hidden layers $\layerNN$, the number of neurons $\neuron$ in each hidden layer, the learning rate $\alpha$ for the model, and the activation function for the hidden and output layers. For the combined architecture and feature search, $\Omega$ additionally includes all pairwise or groupwise-task features mentioned in \cref{subsec:pairwise_affinity} and \cref{subsec:group_affinity}.

\begin{wrapfigure}{L}{0.5\textwidth}
    \begin{minipage}{0.5\textwidth}
    \vspace*{-.8cm}
\begin{algorithm}[H]
\caption{\textbf{NAS Performance Evaluation}($\arch$)}
\label{algo:NAS}
\textbf{Input}: $\arch \gets$ \text{start architecture}, $iter^{max}$, $P$

\begin{algorithmic}[1] %
 
\STATE $\score =$ \textbf{ArchEvaluate}~($\arch$)

 \FOR{$iter = 0, 1, \cdots, iter^{max}$ }
 \STATE $\arch^\prime \gets $ \textbf{RandomNeighbour} ($\arch $)

   \STATE $ \score^\prime \gets$ \textbf{ArchEvaluate}($\arch^\prime$)
   \STATE $\pi^{accept} \gets exp\left((\score- \score^\prime) \cdot P\right)$

    \IF{~\textbf{RandomBoolean}($\pi^{accept}$)}
        \STATE $\arch  \gets \arch^\prime$
        \STATE $\score \gets  \score^\prime$ 
    \ENDIF
 \STATE  $iter \gets iter + 1$
   \ENDFOR

\STATE \textbf{return} $\arch^{best}$: best architecture found
  \end{algorithmic}
\end{algorithm}
    \vspace*{-0.5cm}
\end{minipage}
  \end{wrapfigure}
\textbf{Search Strategy : }We employ a random localized search algorithm for our Neural Architecture Search (NAS) to find the optimal hyper-parameters for the task-affinity predictor network (\cref{algo:randomNeighbour}). At each iteration, the algorithm randomly selects a hyper-parameter, $hp$, to tune using a predefined probability distribution, $\probabilityDistribution_\hyperParam$. We assigned lower probabilities for adding or removing hidden layers to avoid large jumps in the search space. 
When modifying the number of hidden layers, \layerNN, the algorithm can either add or remove a layer and uniformly sample the number of neurons in the new layer. To modify the number of neurons, \neuron, the algorithm randomly selects a layer from the existing layers and changes the number of neurons in that layer by a percentage, $\beta_\neuron$. The learning rate is also modified by a change percentage, $\beta_\alpha$. The activation in the hidden layers and output layers can also be tweaked for optimization. Additionally, we can modify input features, $\featureNN$, by adding or removing attributes from the current model with a probability.

\textbf{Evaluation Criterion : }For performance evaluation, our randomized local architecture search algorithm considers the performance ($R^2$) on the validation set and the model complexity. The score $\score$ of each architecture is computed by $R^2(\arch) - \omega \cdot \textit{model\_complexity}(\arch)$. Factor $\omega$ is a cost ratio between prediction error and model complexity that practitioners can set based on their requirements. We describe the performance evaluation in \cref{algo:NAS}. The search starts from an initial architecture, $\arch$, with an initial score $\score$. Then, it follows an iterative process. In each iteration, the algorithm picks a new architecture $\arch^\prime$ using a random mutation of the current best architecture $\arch$ and obtains a new score $\score^\prime$. 
If the new score $\score^\prime$ of $\arch^\prime$ is higher than the score $\score$ of $\arch$, then the algorithm always accepts $\arch^\prime$ and $\score^\prime$ as the new solution. Otherwise, the algorithm accepts the architecture with a probability that depends on the difference between the score and an exogenous parameter $P$. After a fixed number of iterations $iter^{max}$, the algorithm terminates and returns the best solution $\arch^{best}$ found up to that~point.

\subsection{Final Architecture for Groupwise MTL-Gain Predictor}
\label{appendix:affinity_pred_architecture}
While selecting the task-affinity predictor's final architecture, we adopt the optimal architecture and input feature set obtained through a comprehensive search combining hyper-parameter tuning and feature selection. We stop the hyper-parameter and feature search before reaching convergence to optimize computational efficiency throughout the process. Each search is performed for a fixed number of iterations (approximately $500$). Then, we select the top five architectures discovered during the search phase and train them using the complete training data from our partition sample. These trained models have been evaluated on a separate hold-out test set. We repeat the training and evaluation process multiple times with the set of selected architectures to obtain multiple solution quality measures. Finally, we choose the architecture and feature set with the best average performance as our final choice for the task-affinity predictor.

\begin{table}[h]
    \centering
        \caption{Final Architectures for Task-Affinity Predictors}
    \label{tab:arch_from_nas}
    \resizebox{\linewidth}{!}{%
    \begin{tabular}{*{7}{|c}|}
    \hline
     \textbf{MTL Models} & \textbf{Benchmark} & \textbf{\#Layers} & \textbf{Neurons} & \textbf{Activation} & \textbf{Activation} & \textbf{Learning} \\ 
       &  &  &  & \textbf{(hidden)} & \textbf{(output)} & \textbf{Rate}, $\alpha$ \\ \hline

     & School & 4 & $[88, 30, 44, 90]$ & tanh & linear & $0.001234$ \\ \cline{3-7}
     
     \textbf{Neural Network } & Chemical & 4 & $[14, 10, 21, 15]$ & tanh & linear & $0.001247$ \\ \cline{3-7}
      
      & Landmine & 5 & $[20, 10, 15, 18, 31]$ & ReLU & linear & $0.001524$ \\ \cline{3-7}

       & Parkinson & 4 & $[39, 14, 57, 40]$ & tanh & tanh & $0.001524$ \\ \hline \hline

     & School & 2 & $[10, 17]$ & ReLU & linear & $0.001524$ \\ \cline{3-7}
     
     \textbf{Support Vector} & Chemical & 2 & $[58, 109]$ & ReLU & linear & $0.001508$ \\ \cline{3-7}
      
      \textbf{Machine} & Landmine & 2 & $[26, 30]$ & ReLU & linear & $0.001247$ \\ \cline{3-7}

       & Parkinson & 2 & $[123, 67]$ & ReLU & linear & $0.001234$ \\ \hline \hline

     & School & 2 & $[143, 25]$ & ReLU & linear & $0.001385$ \\ \cline{3-7}
 
 \textbf{Extreme Gradient} & Chemical & 6 & $[13, 10, 8, 29, 19, 15]$ & ReLU & sigmoid & $0.001122$ \\ \cline{3-7}
  
  \textbf{Boost} & Landmine & 2 & $[35, 12]$ & ReLU & linear & $0.001385$ \\ \cline{3-7}

   & Parkinson & 1 & $[150]$ & ReLU & linear & $0.001676$ \\ \hline 
       
    \end{tabular}
    }
\end{table}

\begin{table}[h]
    \centering
    \caption{Final Input Features for Task-Affinity Predictors}
    \label{tab:feature_from_nas}
    \resizebox{\linewidth}{!}{%
    \begin{tabular}{|c|c|c|}
    \hline
     \textbf{MTL Models} & \textbf{Benchmark} & \textbf{Groupwise-task Features (for Input)} \\ \hline

     &  &  
     Pairwise MTL gain $(\mu)$, 
     \#Samples, 
     Pairwise MTL gain $(\sigma^2)$, 
      
    \\ 
      & School &  
     Group-target $\sigma^2$ $(\mu)$, 
     \#Tasks, 
     Group distance ($\mu$), 
     Pairwise $\mathcal{W}_{t_i}\cdot\mathcal{W}_{t_j}$ $(\mu)$\\ \cline{2-3}
     
     \textbf{Neural} & Chemical & 
     Pairwise MTL gain $(\mu)$, Group-target $\sigma$ $(\mu)$, Pairwise MTL gain $(\sigma^2)$\\ \cline{2-3}
      
      \textbf{ Network } & Landmine & Group-target $\sigma^2$ $(\mu)$, Pairwise MTL gain $(\mu)$, Pairwise MTL gain $(\sigma)$\\ \cline{2-3}

       & Parkinson & 
       Pairwise MTL gain $(\mu)$, Pairwise MTL gain $(\sigma)$,
       \\    
       &  & 
        Group-target $\sigma^2$ $(\mu)$, Group distance ($\mu$), Group-target $\sigma$ $(\mu)$
       \\ 
       \hline \hline 

         & School &  
     Pairwise MTL gain $(\mu)$, 
     Pairwise MTL gain $(\sigma)$, \\ \cline{2-3}
     
     \textbf{Support Vector} & Chemical & 
       Pairwise MTL gain $(\mu)$, 
     Pairwise MTL gain $(\sigma)$, \\ \cline{2-3}
      
      \textbf{Machine} & Landmine &      Pairwise MTL gain $(\mu)$, 
     Pairwise MTL gain $(\sigma^2)$, \\ \cline{2-3}

       & Parkinson & 
       Pairwise MTL gain $(\mu)$, Pairwise MTL gain $(\sigma^2)$
       \\ 
       \hline \hline 

         & School &  
     Pairwise MTL gain $(\mu)$, 
     Pairwise MTL gain $(\sigma^2)$,  
     \#Tasks, Group-target $\sigma$ $(\mu)$  \\ \cline{2-3}
     
     \textbf{Extreme} & Chemical & 
     Pairwise MTL gain $(\mu)$, 
     Pairwise MTL gain $(\sigma)$, \#Tasks\\ 
     
     \textbf{Gradient} &  & 
     \#Tasks, 
     Pairwise MTL gain $(\sigma^2)$\\ \cline{2-3}

      \textbf{Boost} & Landmine & 
       Pairwise MTL gain $(\mu)$,
       Group-target $\sigma^2$ $(\mu)$, 
       \#Tasks, 
      \\ 
       &  & 
       Pairwise MTL gain $(\sigma^2)$, 
       Pairwise MTL gain $(\sigma)$, 
       Group-target $\sigma$ $(\mu)$
      \\ \cline{2-3}

       & Parkinson & 
       Pairwise MTL gain $(\mu)$, Pairwise MTL gain $(\sigma)$,
       Group-target $\sigma$ $(\mu)$
       \\ 
       \hline 
    \end{tabular}
    }
\end{table}

\cref{tab:arch_from_nas} and \cref{tab:feature_from_nas} outline the final architectures and input features for the task-affinity predictors for each benchmark, considering various training data gathered using different configurations of the MTL models. In addition to the hyper-parameters mentioned in \cref{tab:arch_from_nas}, we use a varying batch size strategy based on the training sample size. Each predictor model is trained until convergence, typically reaching $150-200$ iterations.

\section{Study of Task Affinity}
In this section, we present additional results for our pairwise task-affinity (\cref{subsec:pairwise_affinity}) and groupwise-task affinity (\cref{subsec:group_affinity}) studies.

\subsection{Pearson Correlation for Features from Pairwise-affinity}
\label{appendix:PearsonCorr_Pairwise}

\textbf{Pearson Correlation for Pairwise-Task Features}

\cref{fig:pearsonCorr_pair,fig:pearsonCorr_pair_svm,fig:pearsonCorr_pair_xgb} show 9 pairwise-task features that correlate the most with the relative MTL gain on each benchmark. Due to space constraints, we shorten the names of some features. The suffixes `nP' and `nS' stand for normalized by product and normalized by sum, respectively. We describe the features in \cref{appendix:pairwise_feature_desc}.

\pgfplotstableread[col sep=comma]{data/Pairwise_PearsonCorrelation_School_NN_top.csv}\PearsonSch
\pgfplotstableread[col sep=comma]{data/Pairwise_PearsonCorrelation_Chemical_NN_top.csv}\PearsonChem
\pgfplotstableread[col sep=comma]{data/Pairwise_PearsonCorrelation_Landmine_NN_top.csv}\PearsonLM
\pgfplotstableread[col sep=comma]{data/Pairwise_PearsonCorrelation_Parkinsons_NN_top.csv}\PearsonPK

\begin{figure}[h]

    \begin{subfigure}[b]{0.2625\linewidth}
    \begin{tikzpicture}[]
            \begin{axis}
            [
                MediumBarPlot,
                width = \linewidth,
                height = 3.5cm,
                ylabel = Pearson Correlation,
                label style = {align = center, font = \small},
                grid=major,
                yticklabel=\pgfmathprintnumber{\tick},
                xtick = data,
                ymin = -1, ymax = 1,
                xmin =-0.425,xmax = 8.3,
                xticklabels from table={\PearsonSch}{Feature},
              x tick label style={font = \tiny,rotate = 90, anchor=east},
                legend pos=north east,font = \scriptsize,
              legend style={at={(0,1)}, anchor=south east}
            ]
            \addplot [PurpleBars] table [y={Pearson_Corr}, x expr=\coordindex] {\PearsonSch};
            
            \end{axis}
        \end{tikzpicture}
        \caption{\textbf{School}}
        \label{subfig:Pearsonpair_sch}
    \end{subfigure}
    \begin{subfigure}[b]{0.2375\linewidth}
        \begin{tikzpicture}[]
                \begin{axis}
                [
                    MediumBarPlot,
                    width = \linewidth,
                    height = 3.5cm,
                    ylabel = Pearson Correlation,
                    label style = {align = center, font = \small},
                    grid=major,
                    yticklabel=\pgfmathprintnumber{\tick},
                    xtick = data,
                    ymin = -1, ymax = 1,
                    xmin =-0.425,xmax = 8.3,
                    xticklabels from table={\PearsonChem}{Feature},
                  x tick label style={font = \tiny,rotate = 90, anchor=east},
                    legend pos=north east,font = \scriptsize,
                  legend style={at={(0,1)}, anchor=south east}
                ]
                \addplot [PurpleBars] table [y={Pearson_Corr}, x expr=\coordindex] {\PearsonChem};

                \end{axis}
            \end{tikzpicture}
            \caption{\textbf{Chemical}}
            \label{subfig:Pearsonpair_chem}
        \end{subfigure}
        \begin{subfigure}[b]{0.245\linewidth}
        \begin{tikzpicture}[]
                \begin{axis}
                [
                    MediumBarPlot,
                    width = \linewidth,
                    height = 3.5cm,
                    ylabel = Pearson Correlation,
                    label style = {align = center, font = \small},
                    grid=major,
                    yticklabel=\pgfmathprintnumber{\tick},
                    xtick = data,
                    ymin = -1, ymax = 1,
                    xmin =-0.425,xmax = 8.3,
                    xticklabels from table={\PearsonLM}{Feature},
                  x tick label style={font = \tiny,rotate = 90, anchor=east},
                    legend pos=north east,font = \scriptsize,
                  legend style={at={(0,1)}, anchor=south east}
                ]
                \addplot [PurpleBars] table [y={Pearson_Corr}, x expr=\coordindex] {\PearsonLM};
                
                \end{axis}
            \end{tikzpicture}
            \caption{\textbf{Landmine}}
            \label{subfig:Pearsonpair_lm}
        \end{subfigure}
         \begin{subfigure}[b]{0.24\linewidth}
        \begin{tikzpicture}[]
                \begin{axis}
                [
                    MediumBarPlot,
                    width = \linewidth,
                    height = 3.5cm,
                    ylabel = Pearson Correlation,
                    label style = {align = center, font = \small},
                    grid=major,
                    yticklabel=\pgfmathprintnumber{\tick},
                    xtick = data,
                    ymin = -1, ymax = 1,
                    xmin =-0.425,xmax = 8.3,
                    xticklabels from table={\PearsonPK}{Feature},
                  x tick label style={font = \tiny,rotate = 90, anchor=east},
                    legend pos=north east,font = \scriptsize,
                  legend style={at={(0,1)}, anchor=south east}
                ]
                \addplot [PurpleBars] table [y={Pearson_Corr}, x expr=\coordindex] {\PearsonPK};
                
                \end{axis}
            \end{tikzpicture}
            \caption{\textbf{Parkinson}}
            \label{subfig:Pearsonpair_pk}
        \end{subfigure}
        \vspace{-1.5em}
\caption{Features that have high Pearson correlation coefficients with relative pairwise MTL gain (sorted by absolute value) - pairwise MTL models are implemented using \textbf{Neural Networks}.}
\label{fig:pearsonCorr_pair}
\vspace{-1.25em}
\end{figure}
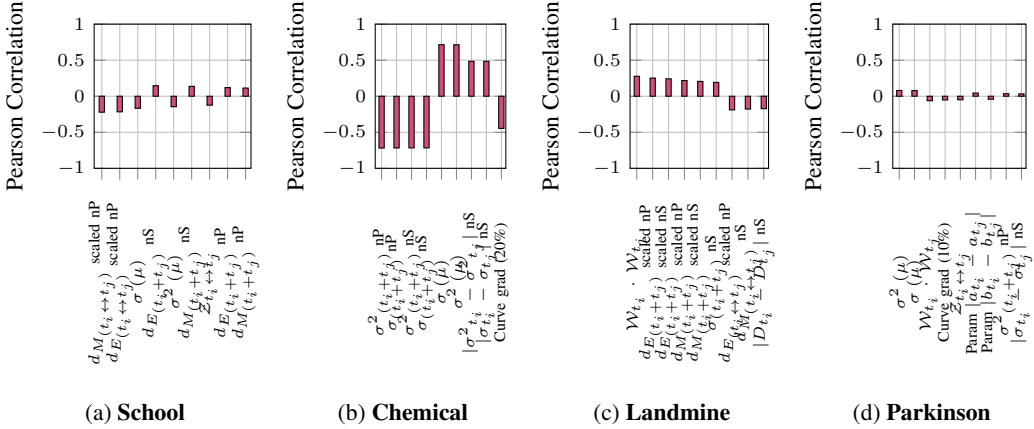

\pgfplotstableread[col sep=comma]{data/Pairwise_PearsonCorrelation_School_SVM_top.csv}\PearsonSch
\pgfplotstableread[col sep=comma]{data/Pairwise_PearsonCorrelation_Chemical_SVM_top.csv}\PearsonChem
\pgfplotstableread[col sep=comma]{data/Pairwise_PearsonCorrelation_Landmine_SVM_top.csv}\PearsonLM
\pgfplotstableread[col sep=comma]{data/Pairwise_PearsonCorrelation_Parkinsons_SVM_top.csv}\PearsonPK

\begin{figure}[h]

    \begin{subfigure}[b]{0.25\linewidth}
    \begin{tikzpicture}[]
            \begin{axis}
            [
                MediumBarPlot,
                width = \linewidth,
                height = 3.5cm,
                ylabel = Pearson Correlation,
                label style = {align = center, font = \small},
                grid=major,
                yticklabel=\pgfmathprintnumber{\tick},
                xtick = data,
                ymin = -1, ymax = 1,
                 xmin =-0.425,xmax = 8.3,
                xticklabels from table={\PearsonSch}{Feature},
              x tick label style={font = \tiny,rotate = 90, anchor=east},
                legend pos=north east,font = \scriptsize,
              legend style={at={(0,1)}, anchor=south east}
            ]
            \addplot [PurpleBars] table [y={Pearson_Corr}, x expr=\coordindex] {\PearsonSch};
            
            \end{axis}
        \end{tikzpicture}
        \caption{\textbf{School}}
        \label{subfig:svm_Pearsonpair_sch}
    \end{subfigure}
    \begin{subfigure}[b]{0.24\linewidth}
        \begin{tikzpicture}[]
                \begin{axis}
                [
                    MediumBarPlot,
                    width = \linewidth,
                    height = 3.5cm,
                    ylabel = Pearson Correlation,
                    label style = {align = center, font = \small},
                    grid=major,
                    yticklabel=\pgfmathprintnumber{\tick},
                    xtick = data,
                    ymin = -1, ymax = 1,
                    xmin =-0.425,xmax = 8.3,
                    xticklabels from table={\PearsonChem}{Feature},
                  x tick label style={font = \tiny,rotate = 90, anchor=east},
                    legend pos=north east,font = \scriptsize,
                  legend style={at={(0,1)}, anchor=south east}
                ]
                \addplot [PurpleBars] table [y={Pearson_Corr}, x expr=\coordindex] {\PearsonChem};

                \end{axis}
            \end{tikzpicture}
            \caption{\textbf{Chemical}}
            \label{subfig:svm_Pearsonpair_chem}
        \end{subfigure}
        \begin{subfigure}[b]{0.25\linewidth}
        \begin{tikzpicture}[]
                \begin{axis}
                [
                    MediumBarPlot,
                    width = \linewidth,
                    height = 3.5cm,
                    ylabel = Pearson Correlation,
                    label style = {align = center, font = \small},
                    grid=major,
                    yticklabel=\pgfmathprintnumber{\tick},
                    xtick = data,
                    ymin = -1, ymax = 1,
                     xmin =-0.425,xmax = 8.3,
                    xticklabels from table={\PearsonLM}{Feature},
                  x tick label style={font = \tiny,rotate = 90, anchor=east},
                    legend pos=north east,font = \scriptsize,
                  legend style={at={(0,1)}, anchor=south east}
                ]
                \addplot [PurpleBars] table [y={Pearson_Corr}, x expr=\coordindex] {\PearsonLM};
                
                \end{axis}
            \end{tikzpicture}
            \caption{\textbf{Landmine}}
            \label{subfig:svm_Pearsonpair_lm}
        \end{subfigure}
         \begin{subfigure}[b]{0.24\linewidth}
        \begin{tikzpicture}[]
                \begin{axis}
                [
                    MediumBarPlot,
                    width = \linewidth,
                    height = 3.5cm,
                    ylabel = Pearson Correlation,
                    label style = {align = center, font = \small},
                    grid=major,
                    yticklabel=\pgfmathprintnumber{\tick},
                    xtick = data,
                    ymin = -1, ymax = 1,
                     xmin =-0.425,xmax = 8.3,
                    xticklabels from table={\PearsonPK}{Feature},
                  x tick label style={font = \tiny,rotate = 90, anchor=east},
                    legend pos=north east,font = \scriptsize,
                  legend style={at={(0,1)}, anchor=south east}
                ]
                \addplot [PurpleBars] table [y={Pearson_Corr}, x expr=\coordindex] {\PearsonPK};
                
                \end{axis}
            \end{tikzpicture}
            \caption{\textbf{Parkinson}}
            \label{subfig:svm_Pearsonpair_pk}
        \end{subfigure}
        \vspace{-1.5em}
\caption{Features that have high Pearson correlation coefficients with relative pairwise MTL gain (sorted by absolute value) - pairwise MTL models are implemented using \textbf{Support Vector Machines (SVM)}.}
\label{fig:pearsonCorr_pair_svm}
\vspace{-0.5em}
\end{figure}
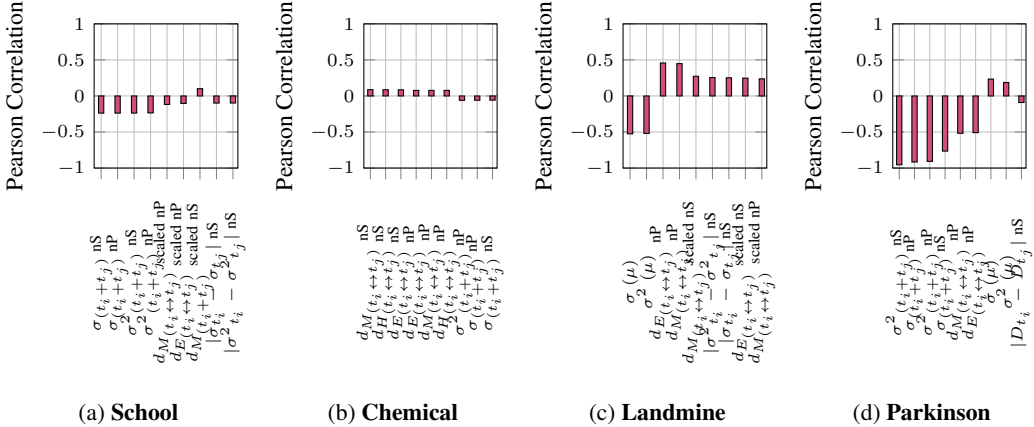

\pgfplotstableread[col sep=comma]{data/Pairwise_PearsonCorrelation_School_xgBoost_top.csv}\PearsonSch
\pgfplotstableread[col sep=comma]{data/Pairwise_PearsonCorrelation_Chemical_xgBoost_top.csv}\PearsonChem
\pgfplotstableread[col sep=comma]{data/Pairwise_PearsonCorrelation_Landmine_xgBoost_top.csv}\PearsonLM
\pgfplotstableread[col sep=comma]{data/Pairwise_PearsonCorrelation_Parkinsons_xgBoost_top.csv}\PearsonPK
\begin{figure}[h!]

    \begin{subfigure}[b]{0.25\linewidth}
    \begin{tikzpicture}[]
            \begin{axis}
            [
                MediumBarPlot,
                width = \linewidth,
                height = 3.5cm,
                ylabel = Pearson Correlation,
                label style = {align = center, font = \small},
                grid=major,
                yticklabel=\pgfmathprintnumber{\tick},
                xtick = data,
                ymin = -1, ymax = 1,
                 xmin =-0.425,xmax = 8.3,
                xticklabels from table={\PearsonSch}{Feature},
              x tick label style={font = \tiny,rotate = 90, anchor=east},
                legend pos=north east,font = \scriptsize,
              legend style={at={(0,1)}, anchor=south east}
            ]
            \addplot [PurpleBars] table [y={Pearson_Corr}, x expr=\coordindex] {\PearsonSch};
            \end{axis}
        \end{tikzpicture}
        \caption{\textbf{School}}
        \label{subfig:xgb_Pearsonpair_sch}
    \end{subfigure}
    \begin{subfigure}[b]{0.24\linewidth}
        \begin{tikzpicture}[]
                \begin{axis}
                [
                    MediumBarPlot,
                    width = \linewidth,
                    height = 3.5cm,
                    ylabel = Pearson Correlation,
                    label style = {align = center, font = \small},
                    grid=major,
                    yticklabel=\pgfmathprintnumber{\tick},
                    xtick = data,
                    ymin = -1, ymax = 1,
                     xmin =-0.425,xmax = 8.3,
                    xticklabels from table={\PearsonChem}{Feature},
                  x tick label style={font = \tiny,rotate = 90, anchor=east},
                    legend pos=north east,font = \scriptsize,
                  legend style={at={(0,1)}, anchor=south east}
                ]
                \addplot [PurpleBars] table [y={Pearson_Corr}, x expr=\coordindex] {\PearsonChem};

                \end{axis}
            \end{tikzpicture}
            \caption{\textbf{Chemical}}
            \label{subfig:xgb_Pearsonpair_chem}
        \end{subfigure}
        \begin{subfigure}[b]{0.25\linewidth}
        \begin{tikzpicture}[]
                \begin{axis}
                [
                    MediumBarPlot,
                    width = \linewidth,
                    height = 3.5cm,
                    ylabel = Pearson Correlation,
                    label style = {align = center, font = \small},
                    grid=major,
                    yticklabel=\pgfmathprintnumber{\tick},
                    xtick = data,
                    ymin = -1, ymax = 1,
                     xmin =-0.425,xmax = 8.3,
                    xticklabels from table={\PearsonLM}{Feature},
                  x tick label style={font = \tiny,rotate = 90, anchor=east},
                    legend pos=north east,font = \scriptsize,
                  legend style={at={(0,1)}, anchor=south east}
                ]
                \addplot [PurpleBars] table [y={Pearson_Corr}, x expr=\coordindex] {\PearsonLM};
                
                \end{axis}
            \end{tikzpicture}
            \caption{\textbf{Landmine}}
            \label{subfig:xgb_Pearsonpair_lm}
        \end{subfigure}
         \begin{subfigure}[b]{0.245\linewidth}
        \begin{tikzpicture}[]
                \begin{axis}
                [
                    MediumBarPlot,
                    width = \linewidth,
                    height = 3.5cm,
                    ylabel = Pearson Correlation,
                    label style = {align = center, font = \small},
                    grid=major,
                    yticklabel=\pgfmathprintnumber{\tick},
                    xtick = data,
                    ymin = -1, ymax = 1,
                     xmin =-0.425,xmax = 8.3,
                    xticklabels from table={\PearsonPK}{Feature},
                  x tick label style={font = \tiny,rotate = 90, anchor=east},
                    legend pos=north east,font = \scriptsize,
                  legend style={at={(0,1)}, anchor=south east}
                ]
                \addplot [PurpleBars] table [y={Pearson_Corr}, x expr=\coordindex] {\PearsonPK};
                
                \end{axis}
            \end{tikzpicture}
            \caption{\textbf{Parkinson}}
            \label{subfig:xgb_Pearsonpair_pk}
        \end{subfigure}
        \vspace{-1.5em}
\caption{Features that have high Pearson correlation coefficients with relative pairwise MTL gain (sorted by absolute value) - pairwise MTL models are implemented using \textbf{Extreme Gradient Boosting trees (XGBoost)}.}
\label{fig:pearsonCorr_pair_xgb}
\vspace{-0.5em}
\end{figure}
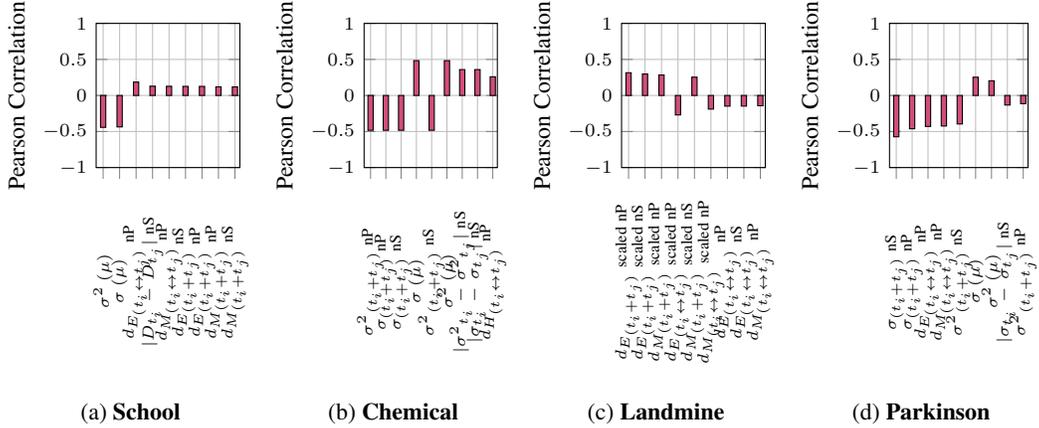

\cref{fig:pearsonCorr_pair}, \cref{fig:pearsonCorr_pair_svm}, and \cref{fig:pearsonCorr_pair_xgb} use the relative pairwise MTL gain values when two tasks are trained with MTLs using neural networks, support vector machines (SVM), and extreme gradient boosting (XGBoost) trees, respectively.
From these figures, we observe that features representing the distance between the samples of the tasks and the variance (or std.\ dev.) of the target attribute in the combined sample are highly correlated with relative MTL gain. Furthermore, irrespective of the MTL technique, the highly correlated features remain almost consistent for the same benchmark.

\textbf{Pearson Correlation for Single-Task Features}
\pgfplotstableread[col sep=comma]{data/PearsonCorrelation_TaskSpecific_NN.csv}\pearsonPairTS

\pgfplotstableread[col sep=comma]{data/PearsonCorrelation_TaskSpecific_SVM.csv}\pearsonPairTSSVM

\pgfplotstableread[col sep=comma]{data/PearsonCorrelation_TaskSpecific_xgBoost.csv}\pearsonPairTSxgb

\begin{figure}[h!]
    \centering
    \begin{subfigure}[b]{\linewidth}
        \begin{tikzpicture}[]
            \begin{axis}
            [
                MediumBarPlot,
                width = \linewidth,
                height = 3.5cm,
                xlabel=\textbf{Features specific to tasks and STL using Neural Network},
                ylabel = Pearson Correlation,
                label style = {align = center, font=\small},
                grid=major,
                ymin = -1, ymax = 1,
                xmin = -1,xmax = 15.75,
                yticklabel=\pgfmathprintnumber{\tick},
                xtick = data,
                xticklabels from table={\pearsonPairTS}{Feature},
               x tick label style={rotate=90,font = \scriptsize, anchor=east},
               legend columns = 5,
                legend pos=north east,font = \scriptsize,
                legend style={at={(0,0)}, anchor=south west}
            ]
            \addplot [PurpleBars] table [y={Avg}, x expr=\coordindex] {\pearsonPairTS};
            \addlegendentry{Avg};
            
            \addplot[only marks, mark=+, thick,draw=black!50]
            table [y={School}, x expr=\coordindex] {\pearsonPairTS};
            \addlegendentry{School};
            
            \addplot[only marks, mark=o, thick,draw=black!50]
            table [y={Chemical}, x expr=\coordindex] {\pearsonPairTS};
                \addlegendentry{Chemical};
                
             \addplot[only marks, mark=*, thick,draw=black!50]
            table [y={Landmine}, x expr=\coordindex] {\pearsonPairTS};
            \addlegendentry{Landmine};

            \addplot[only marks, mark=pentagon, thick,draw=black!50]
            table [y={Parkinsons}, x expr=\coordindex] {\pearsonPairTS};
            \addlegendentry{Parkinson};
            \end{axis}
            \vspace{-1.5em}
        \end{tikzpicture}
        \end{subfigure}
        \begin{subfigure}[b]{\linewidth}
            \begin{tikzpicture}[]
            \begin{axis}
            [
                MediumBarPlot,
                width = \linewidth,
                height = 3.5cm,
                xlabel=\textbf{Features specific to tasks and STL using Support Vector Machine},
                ylabel = Pearson Correlation,
                label style = {align = center, font=\small},
                grid=major,
                ymin = -1, ymax = 1,
                xmin = -1,xmax = 4.5,
                yticklabel=\pgfmathprintnumber{\tick},
                xtick = data,
                xticklabels from table={\pearsonPairTSSVM}{Feature},
               x tick label style={font = \scriptsize},
               legend columns = 5,
               legend pos=north east,font = \scriptsize,
                legend style={at={(0,0)}, anchor=south west}
            ]
            \addplot [PurpleBars] table [y={Avg}, x expr=\coordindex] {\pearsonPairTSSVM};
            \addlegendentry{Avg};
            
            \addplot[only marks, mark=+, thick,draw=black!50]
            table [y={School}, x expr=\coordindex] {\pearsonPairTSSVM};
            \addlegendentry{School};
            
            \addplot[only marks, mark=o, thick,draw=black!50]
            table [y={Chemical}, x expr=\coordindex] {\pearsonPairTSSVM};
                \addlegendentry{Chemical};
                
             \addplot[only marks, mark=*, thick,draw=black!50]
            table [y={Landmine}, x expr=\coordindex] {\pearsonPairTSSVM};
            \addlegendentry{Landmine};

            \addplot[only marks, mark=pentagon, thick,draw=black!50]
            table [y={Parkinsons}, x expr=\coordindex] {\pearsonPairTSSVM};
            \addlegendentry{Parkinson};
            \end{axis}
            \label{subfig:pearsonCorr_pair_taskSpecific_svm}
        \end{tikzpicture}
        \end{subfigure}
        \begin{subfigure}[b]{\linewidth}
            \begin{tikzpicture}[]
            \begin{axis}
            [
                MediumBarPlot,
                width = \linewidth,
                height = 3.5cm,
                xlabel=\textbf{Features specific to tasks and STL using XGBoost},
                ylabel = Pearson Correlation,
                label style = {align = center, font=\small},
                grid=major,
                ymin = -1, ymax = 1,
                xmin = -1,xmax = 4.5,
                yticklabel=\pgfmathprintnumber{\tick},
                xtick = data,
                xticklabels from table={\pearsonPairTSxgb}{Feature},
               x tick label style={font = \scriptsize},
               legend columns = 5,
                legend pos=north east,font = \scriptsize,
                legend style={at={(0,0)}, anchor=south west}
            ]
            \addplot [PurpleBars] table [y={Avg}, x expr=\coordindex] {\pearsonPairTSxgb};
            \addlegendentry{Avg};
            
            \addplot[only marks, mark=+, thick,draw=black!50]
            table [y={School}, x expr=\coordindex] {\pearsonPairTSxgb};
            \addlegendentry{School};
            
            \addplot[only marks, mark=o, thick,draw=black!50]
            table [y={Chemical}, x expr=\coordindex] {\pearsonPairTSxgb};
                \addlegendentry{Chemical};
                
             \addplot[only marks, mark=*, thick,draw=black!50]
            table [y={Landmine}, x expr=\coordindex] {\pearsonPairTSxgb};
            \addlegendentry{Landmine};

            \addplot[only marks, mark=pentagon, thick,draw=black!50]
            table [y={Parkinsons}, x expr=\coordindex] {\pearsonPairTSxgb};
            \addlegendentry{Parkinson};
            \end{axis}
        \label{subfig:pearsonCorr_pair_taskSpecific_xgb}
        \end{tikzpicture}
        \end{subfigure}
        
\caption{Pearson correlation coefficient of each single-task feature with relative pairwise MTL gain (sorted by absolute correlation value) when two tasks are trained together using \textbf{Neural Networks}, \textbf{Support Vector Machines (SVM)}, and \textbf{Extreme Gradient Boosting (XGBoost)} trees, respectively.}
\label{fig:pearsonCorr_pair_taskSpecific}
\vspace{-0.5em}
\end{figure}

\cref{fig:pearsonCorr_pair_taskSpecific} shows the Pearson correlation between each single-task feature and the relative gain from pairwise MTL training when MTL models are employed using neural networks, support vector machines 
(SVM), and extreme gradient boosting (XGBoost) trees, respectively. We describe all the features in \cref{appendix:pairwise_feature_desc}. As mentioned in \cref{appendix:pairwise_feature_desc}, we could not consider some STL-based features for MTL implementations with SVM and XGBoost.

From \cref{fig:pearsonCorr_pair_taskSpecific}, we observe that features with high correlation include the log-curve fitted parameters of neural-network-based STL, sample size, and the standard deviation of the target attribute. \cref{subfig:pearsonCorr_pair_taskSpecific_svm,subfig:pearsonCorr_pair_taskSpecific_xgb} show that the most correlated single task features are the dataset size of each task and the individual STL-based loss.

\subsection{Usefulness of Pairwise-Task Features}
\label{appendix:Predictor_Usefulness_TR}
\pgfplotstableread[col sep=comma]{data/Pairwise_Individual_Usefulness_avg_NN_all.csv}\nnPairwiseTR
\pgfplotstableread[col sep=comma]{data/Pairwise_Individual_Usefulness_avg_SVM_all.csv}\svmPairwiseTR
\pgfplotstableread[col sep=comma]{data/Pairwise_Individual_Usefulness_avg_xgBoost_all.csv}\xgbPairwiseTR

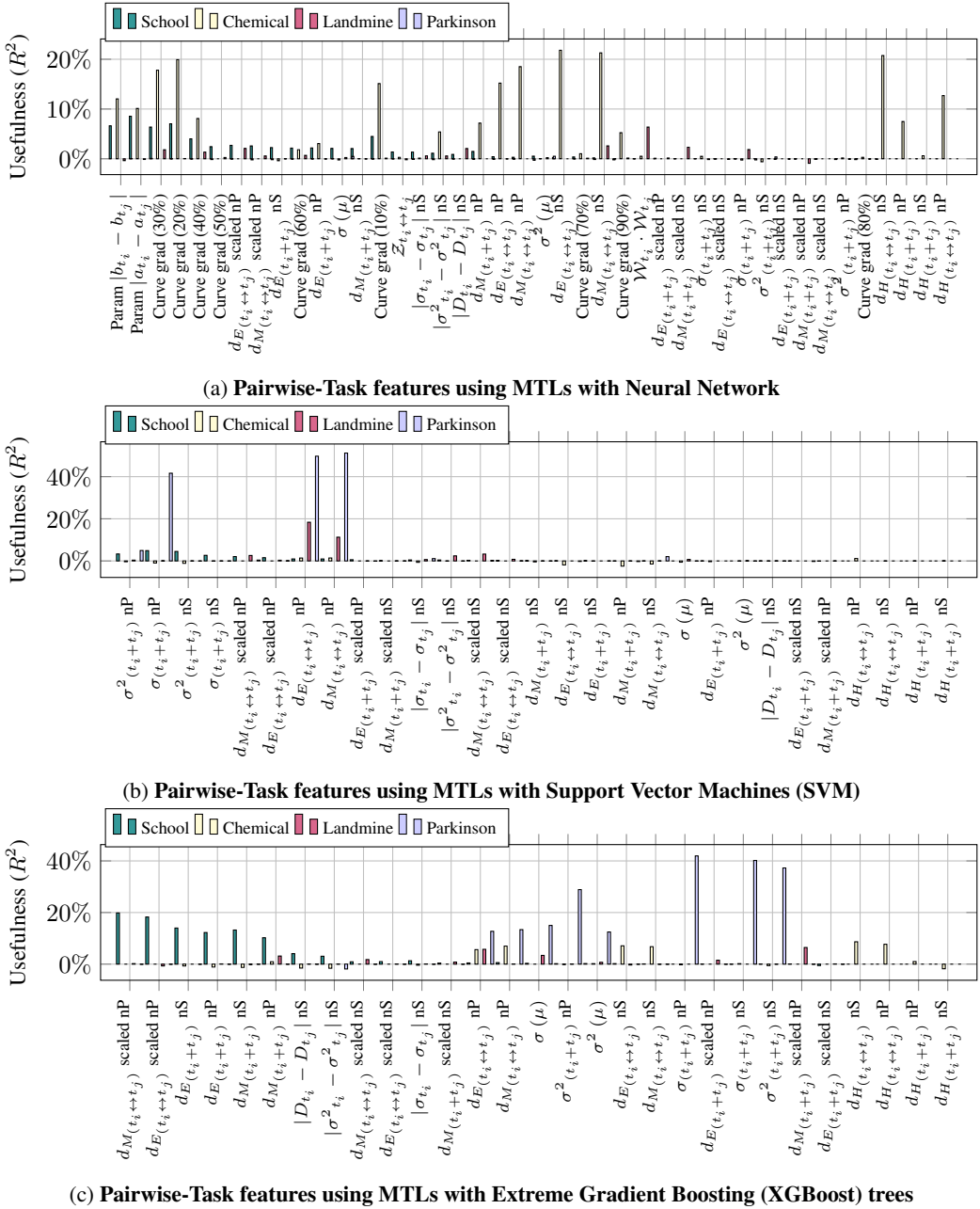
\begin{figure}[h!]{}
    \begin{subfigure}[b]{\linewidth}
    \begin{tikzpicture}[
        /pgf/declare function={
        BarWidth = 0.1;
        BarShift = BarWidth/2 + 0.08;
    },
    ]
            \begin{axis}
            [
            /pgf/bar width=BarWidth,
            /pgf/bar shift=BarShift,
                ybar,
                width = \linewidth,
                height = 3.5cm,
                ylabel = Usefulness (\textit{$R^2$}),
                label style = {align = center, font = \small},
                grid=major,                yticklabel=\pgfmathprintnumber{\tick}{$\%$},
                xtick = data,
                xmin = -1,xmax = 42.5,
                xticklabels from table={\nnPairwiseTR}{Feature},
              x tick label style={font = \scriptsize,rotate = 90, anchor=east},
              legend columns = 4,
                legend style={at={(0.005,0.98)}, anchor=south west,font = \scriptsize}
            ]
            \addplot [draw=black,fill=teal!80] table [y={School}, x expr=\coordindex] {\nnPairwiseTR};
            \addlegendentry{School};

            \addplot [draw=black,fill=yellow!20] table [y={Chemical}, x expr=\coordindex] {\nnPairwiseTR};
            \addlegendentry{Chemical};

            \addplot [draw=black,fill=purple!60] table [y={Landmine}, x expr=\coordindex] {\nnPairwiseTR};
            \addlegendentry{Landmine};

            \addplot [draw=black,fill=blue!20] table [y={Parkinsons}, x expr=\coordindex] {\nnPairwiseTR};
            \addlegendentry{Parkinson};
            
            \end{axis}
        \end{tikzpicture}
        \caption{\textbf{Pairwise-Task features using MTLs with Neural Network}}
        \label{subfig:nn_predictor_pairwise}
        \end{subfigure}
        \begin{subfigure}[b]{\linewidth}
            \begin{tikzpicture}[
        /pgf/declare function={
        BarWidth = 0.1;
        BarShift = BarWidth/2 + 0.08;
    },
    ]
            \begin{axis}
            [
            /pgf/bar width=BarWidth,
            /pgf/bar shift=BarShift,
                ybar,
                width = \linewidth,
                height = 3.5cm,
                ylabel = Usefulness (\textit{$R^2$}),
                label style = {align = center, font = \small},
                grid=major,                yticklabel=\pgfmathprintnumber{\tick}{$\%$},
                xtick = data,
                xmin = -1,xmax = 29,
                xticklabels from table={\svmPairwiseTR}{Feature},
              x tick label style={font = \scriptsize,rotate = 90, anchor=east},
              legend columns = 4,
                legend style={at={(0.005,0.98)}, anchor=south west,font = \scriptsize}
            ]
            \addplot [draw=black,fill=teal!80] table [y={School}, x expr=\coordindex] {\svmPairwiseTR};
            \addlegendentry{School};

            \addplot [draw=black,fill=yellow!20] table [y={Chemical}, x expr=\coordindex] {\svmPairwiseTR};
            \addlegendentry{Chemical};

            \addplot [draw=black,fill=purple!60] table [y={Landmine}, x expr=\coordindex] {\svmPairwiseTR};
            \addlegendentry{Landmine};

            \addplot [draw=black,fill=blue!20] table [y={Parkinsons}, x expr=\coordindex] {\svmPairwiseTR};
            \addlegendentry{Parkinson};
            
            \end{axis}
        \end{tikzpicture}
               \caption{\textbf{Pairwise-Task features using MTLs with Support Vector Machines (SVM)}}
        \label{subfig:svm_predictor_pairwise}
        \end{subfigure}
        \begin{subfigure}[b]{\linewidth}
        \begin{tikzpicture}[
        /pgf/declare function={
        BarWidth = 0.1;
        BarShift = BarWidth/2 + 0.08;
    },
    ]
            \begin{axis}
            [
            /pgf/bar width=BarWidth,
            /pgf/bar shift=BarShift,
                ybar,
                width = \linewidth,
                height = 3.5cm,
                ylabel = Usefulness (\textit{$R^2$}),
                label style = {align = center, font = \small},
                grid=major,                yticklabel=\pgfmathprintnumber{\tick}{$\%$},
                xtick = data,
                xmin = -1,xmax = 29,
                xticklabels from table={\xgbPairwiseTR}{Feature},
              x tick label style={font = \scriptsize,rotate = 90, anchor=east},
              legend columns = 4,
                legend style={at={(0.005,0.98)}, anchor=south west,font = \scriptsize}
            ]
            \addplot [draw=black,fill=teal!80] table [y={School}, x expr=\coordindex] {\xgbPairwiseTR};
            \addlegendentry{School};

            \addplot [draw=black,fill=yellow!20] table [y={Chemical}, x expr=\coordindex] {\xgbPairwiseTR};
            \addlegendentry{Chemical};

            \addplot [draw=black,fill=purple!60] table [y={Landmine}, x expr=\coordindex] {\xgbPairwiseTR};
            \addlegendentry{Landmine};

            \addplot [draw=black,fill=blue!20] table [y={Parkinsons}, x expr=\coordindex] {\xgbPairwiseTR};
            \addlegendentry{Parkinson};
            
            \end{axis}
        \end{tikzpicture}
        \caption{\textbf{Pairwise-Task features using MTLs with Extreme Gradient Boosting (XGBoost) trees}}
        \label{subfig:xgb_predictor_pairwise}
        \end{subfigure}
\caption{The usefulness of each feature (described in \cref{appendix:pairwise_feature_desc}) for predicting relative pairwise MTL gain when two tasks are trained together using \textbf{Neural Networks}, \textbf{Support Vector Machines (SVM)}, and \textbf{Extreme Gradient Boosting (XGBoost) trees}, respectively.}
\label{fig:predictor_pairwise_ALL}
\end{figure}
\vspace{-0.5em}
\cref{fig:predictor_pairwise_ALL} shows the usefulness of each pairwise-task feature for predicting relative MTL gain for pairs of tasks trained using neural networks (NN), support vector machines (SVM), and extreme gradient boosting (XGBoost) trees, respectively. The features are described in detail in \cref{appendix:pairwise_feature_desc}. Due to space constraints, we shorten the names of some features. The suffixes `nP' and `nS' stand for normalized by product and normalized by sum, respectively.

The vertical axes in \cref{subfig:nn_predictor_pairwise,subfig:svm_predictor_pairwise,subfig:xgb_predictor_pairwise} indicate the variance explained ($R^2$-value) by each feature when we fit a model to predict relative MTL gain using only that feature. 
When the underlying STL and MTL models are neural networks, we find that the most useful features are related to STL curve gradients for three of the four benchmarks. For all different MTL implementation (NN, SVM, and XGBoost), features representing the distance between samples and the target attribute's variance (or std. dev.) are generally useful predictors.

\subsection{Pearson Correlation for Groupwise-Task Features}
\label{appendix:groupaffinity_pearson}

\pgfplotstableread[col sep=comma]{data/Groupwise_PearsonCorrelation_School.csv}\PearsonGroupwiseSch
\pgfplotstableread[col sep=comma]{data/Groupwise_PearsonCorrelation_Chemical.csv}\PearsonGroupwiseChem
\pgfplotstableread[col sep=comma]{data/Groupwise_PearsonCorrelation_Landmine.csv}\PearsonGroupwiseLM
\pgfplotstableread[col sep=comma]{data/Groupwise_PearsonCorrelation_Parkinsons.csv}\PearsonGroupwisePK
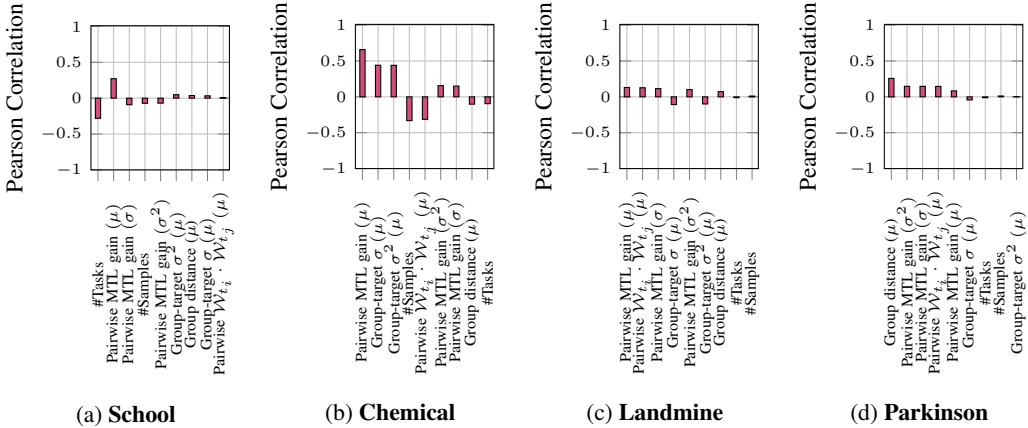
\begin{figure}[h!]
    \begin{subfigure}[b]{0.245\linewidth}
    \begin{tikzpicture}[]
            \begin{axis}
            [
                MediumBarPlot,
                width = \linewidth,
                height = 3.5cm,
                ylabel = Pearson Correlation,
                label style = {align = center, font = \footnotesize},
                grid=major,
                yticklabel=\pgfmathprintnumber{\tick},
                xtick = data,
                ymin = -1, ymax = 1.01,
                xmin = -0.45, xmax = 8.4,
                xticklabels from table={\PearsonGroupwiseSch}{Feature},
              x tick label style={font = \tiny,rotate = 90, anchor=east},
                legend pos=north east,font = \tiny,
              legend style={at={(0,1)}, anchor=south east}
            ]
            \addplot [PurpleBars] table [y={Pearson_Corr}, x expr=\coordindex] {\PearsonGroupwiseSch};
            \end{axis}
        \end{tikzpicture}
        \caption{\textbf{School}}
        \vspace{-0.5pt}
        \label{subfig:PearsonCorr_sch}
    \end{subfigure}
    \begin{subfigure}[b]{0.245\linewidth}
        \begin{tikzpicture}[]
                \begin{axis}
                [
                    MediumBarPlot,
                    width = \linewidth,
                    height = 3.5cm,
                    ylabel = Pearson Correlation,
                    label style = {align = center, font = \footnotesize},
                    grid=major,
                    yticklabel=\pgfmathprintnumber{\tick},
                    xtick = data,
                    ymin = -1, ymax = 1.01,
                    xmin = -0.45, xmax = 8.4,
                    xticklabels from table={\PearsonGroupwiseChem}{Feature},
                  x tick label style={font = \tiny,rotate = 90, anchor=east},
                    legend pos=north east,font = \tiny,
                  legend style={at={(0,1)}, anchor=south east}
                ]
                \addplot [PurpleBars] table [y={Pearson_Corr}, x expr=\coordindex] {\PearsonGroupwiseChem};
                \end{axis}
            \end{tikzpicture}
            \caption{\textbf{Chemical}}
            \label{subfig:PearsonCorr_chem}
        \end{subfigure}
        \begin{subfigure}[b]{0.245\linewidth}
        \begin{tikzpicture}[]
                \begin{axis}
                [
                    MediumBarPlot,
                    width = \linewidth,
                    height = 3.5cm,
                    ylabel = Pearson Correlation,
                    label style = {align = center, font = \footnotesize},
                    grid=major,
                    yticklabel=\pgfmathprintnumber{\tick},
                    xtick = data,
                    ymin = -1,ymax=1.01,
                    xmin = -0.45, xmax = 8.4,
                    xticklabels from table={\PearsonGroupwiseLM}{Feature},
                  x tick label style={font = \tiny,rotate = 90, anchor=east},
                    legend pos=north east,font = \tiny,
                  legend style={at={(0,1)}, anchor=south east}
                ]
                \addplot [PurpleBars] table [y={Pearson_Corr}, x expr=\coordindex] {\PearsonGroupwiseLM};
                \end{axis}
            \end{tikzpicture}
            \caption{\textbf{Landmine}}
            \label{subfig:PearsonCorr_lm}
        \end{subfigure}
        \begin{subfigure}[b]{0.245\linewidth}
        \begin{tikzpicture}[]
                \begin{axis}
                [
                    MediumBarPlot,
                    width = \linewidth,
                    height = 3.5cm,
                    ylabel = Pearson Correlation,
                    label style = {align = center, font = \footnotesize},
                    grid=major,
                    yticklabel=\pgfmathprintnumber{\tick},
                    xtick = data,
                    ymin = -1, ymax = 1.01,
                    xmin = -0.45, xmax = 8.4,
                    xticklabels from table={\PearsonGroupwisePK}{Feature},
                  x tick label style={font = \tiny,rotate = 90, anchor=east},
                    legend pos=north east,font = \tiny,
                  legend style={at={(0,1)}, anchor=south east}
                ]
                \addplot [PurpleBars] table [y={Pearson_Corr}, x expr=\coordindex] {\PearsonGroupwisePK};
                \end{axis}
            \end{tikzpicture}
            \caption{\textbf{Parkinson}}
            \label{subfig:PearsonCorr_pk}
        \end{subfigure}
        \vspace{-0.8em}
\caption{Pearson correlation between groupwise-task features and relative MTL gain for a group of tasks trained using \textbf{Neural Network}. The mean of the relative pairwise MTL gains is highly correlated with the group's relative MTL gain.}
\label{fig:pearsonCorr_tgp}
\end{figure}
\vspace{-0.5em}
\pgfplotstableread[col sep=comma]{data/Groupwise_PearsonCorrelation_School_SVM.csv}\PearsonGroupwiseSch
\pgfplotstableread[col sep=comma]{data/Groupwise_PearsonCorrelation_Chemical_SVM.csv}\PearsonGroupwiseChem
\pgfplotstableread[col sep=comma]{data/Groupwise_PearsonCorrelation_Landmine_SVM.csv}\PearsonGroupwiseLM
\pgfplotstableread[col sep=comma]{data/Groupwise_PearsonCorrelation_Parkinsons_SVM.csv}\PearsonGroupwisePK

\begin{figure}[h!]
    \begin{subfigure}[b]{0.245\linewidth}
    \begin{tikzpicture}[]
            \begin{axis}
            [
                MediumBarPlot,
                width = \linewidth,
                height = 3.5cm,
                ylabel = Pearson Correlation,
                label style = {align = center, font = \footnotesize},
                grid=major,
                yticklabel=\pgfmathprintnumber{\tick},
                xtick = data,
                ymin = -1, ymax = 1.01,
                xmin = -0.45, xmax = 7.4,
                xticklabels from table={\PearsonGroupwiseSch}{Feature},
              x tick label style={font = \tiny,rotate = 90, anchor=east},
                legend pos=north east,font = \tiny,
              legend style={at={(0,1)}, anchor=south east}
            ]
            \addplot [PurpleBars] table [y={Pearson_Corr}, x expr=\coordindex] {\PearsonGroupwiseSch};
            \end{axis}
        \end{tikzpicture}
        \caption{\textbf{School}}
        \vspace{-0.8pt}
        \label{subfig:PearsonCorr_sch_svm}
    \end{subfigure}
    \begin{subfigure}[b]{0.245\linewidth}
        \begin{tikzpicture}[]
                \begin{axis}
                [
                    MediumBarPlot,
                    width = \linewidth,
                    height = 3.5cm,
                    ylabel = Pearson Correlation,
                    label style = {align = center, font = \footnotesize},
                    grid=major,
                    yticklabel=\pgfmathprintnumber{\tick},
                    xtick = data,
                    ymin = -1, ymax = 1.01,
                    xmin = -0.45, xmax = 7.4,
                    xticklabels from table={\PearsonGroupwiseChem}{Feature},
                  x tick label style={font = \tiny,rotate = 90, anchor=east},
                    legend pos=north east,font = \tiny,
                  legend style={at={(0,1)}, anchor=south east}
                ]
                \addplot [PurpleBars] table [y={Pearson_Corr}, x expr=\coordindex] {\PearsonGroupwiseChem};
                \end{axis}
            \end{tikzpicture}
            \caption{\textbf{Chemical}}
            \label{subfig:PearsonCorr_chem_svm}
        \end{subfigure}
        \begin{subfigure}[b]{0.245\linewidth}
        \begin{tikzpicture}[]
                \begin{axis}
                [
                    MediumBarPlot,
                    width = \linewidth,
                    height = 3.5cm,
                    ylabel = Pearson Correlation,
                    label style = {align = center, font = \footnotesize},
                    grid=major,
                    yticklabel=\pgfmathprintnumber{\tick},
                    xtick = data,
                    ymin = -1,ymax=1.01,
                    xmin = -0.45, xmax = 7.4,
                    xticklabels from table={\PearsonGroupwiseLM}{Feature},
                  x tick label style={font = \tiny,rotate = 90, anchor=east},
                    legend pos=north east,font = \tiny,
                  legend style={at={(0,1)}, anchor=south east}
                ]
                \addplot [PurpleBars] table [y={Pearson_Corr}, x expr=\coordindex] {\PearsonGroupwiseLM};
                \end{axis}
            \end{tikzpicture}
            \caption{\textbf{Landmine}}
            \label{subfig:PearsonCorr_lm_svm}
        \end{subfigure}
        \begin{subfigure}[b]{0.245\linewidth}
        \begin{tikzpicture}[]
                \begin{axis}
                [
                    MediumBarPlot,
                    width = \linewidth,
                    height = 3.5cm,
                    ylabel = Pearson Correlation,
                    label style = {align = center, font = \footnotesize},
                    grid=major,
                    yticklabel=\pgfmathprintnumber{\tick},
                    xtick = data,
                    ymin = -1, ymax = 1.01,
                    xmin = -0.45, xmax = 7.4,
                    xticklabels from table={\PearsonGroupwisePK}{Feature},
                  x tick label style={font = \tiny,rotate = 90, anchor=east},
                    legend pos=north east,font = \tiny,
                  legend style={at={(0,1)}, anchor=south east}
                ]
                \addplot [PurpleBars] table [y={Pearson_Corr}, x expr=\coordindex] {\PearsonGroupwisePK};
                \end{axis}
            \end{tikzpicture}
            \caption{\textbf{Parkinson}}
            \label{subfig:PearsonCorr_pk_svm}
        \end{subfigure}
        \vspace{-0.8em}
\caption{Pearson correlation between groupwise-task features and relative MTL gain for a group of tasks trained using \textbf{Support Vector Machine (SVM)}.}
\label{fig:pearsonCorr_tgp_svm}
\end{figure}
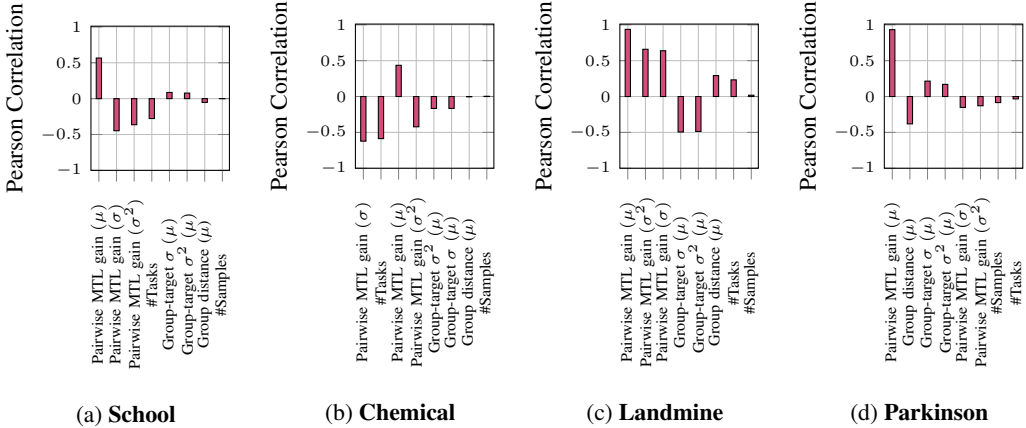
\vspace{-0.5em}
\pgfplotstableread[col sep=comma]{data/Groupwise_PearsonCorrelation_School_xgBoost.csv}\PearsonGroupwiseSch
\pgfplotstableread[col sep=comma]{data/Groupwise_PearsonCorrelation_Chemical_xgBoost.csv}\PearsonGroupwiseChem
\pgfplotstableread[col sep=comma]{data/Groupwise_PearsonCorrelation_Landmine_xgBoost.csv}\PearsonGroupwiseLM
\pgfplotstableread[col sep=comma]{data/Groupwise_PearsonCorrelation_Parkinsons_xgBoost.csv}\PearsonGroupwisePK

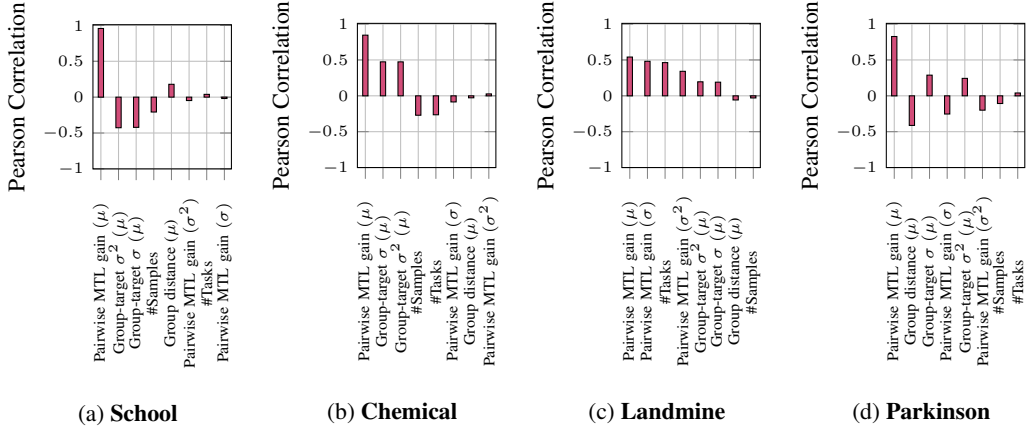
\begin{figure}[h!]
    \begin{subfigure}[b]{0.245\linewidth}
    \begin{tikzpicture}[]
            \begin{axis}
            [
                MediumBarPlot,
                width = \linewidth,
                height = 3.5cm,
                ylabel = Pearson Correlation,
                label style = {align = center, font = \footnotesize},
                grid=major,
                yticklabel=\pgfmathprintnumber{\tick},
                xtick = data,
                ymin = -1, ymax = 1.01,
                xmin = -0.45, xmax = 7.4,
                xticklabels from table={\PearsonGroupwiseSch}{Feature},
              x tick label style={font = \tiny,rotate = 90, anchor=east},
                legend pos=north east,font = \tiny,
              legend style={at={(0,1)}, anchor=south east}
            ]
            \addplot [PurpleBars] table [y={Pearson_Corr}, x expr=\coordindex] {\PearsonGroupwiseSch};
            \end{axis}
        \end{tikzpicture}
        \caption{\textbf{School}}
        \vspace{-0.5pt}
        \label{subfig:PearsonCorr_sch_xgB}
    \end{subfigure}
    \begin{subfigure}[b]{0.245\linewidth}
        \begin{tikzpicture}[]
                \begin{axis}
                [
                    MediumBarPlot,
                    width = \linewidth,
                    height = 3.5cm,
                    ylabel = Pearson Correlation,
                    label style = {align = center, font = \footnotesize},
                    grid=major,
                    yticklabel=\pgfmathprintnumber{\tick},
                    xtick = data,
                    ymin = -1, ymax = 1.01,
                    xmin = -0.45, xmax = 7.4,
                    xticklabels from table={\PearsonGroupwiseChem}{Feature},
                  x tick label style={font = \tiny,rotate = 90, anchor=east},
                    legend pos=north east,font = \tiny,
                  legend style={at={(0,1)}, anchor=south east}
                ]
                \addplot [PurpleBars] table [y={Pearson_Corr}, x expr=\coordindex] {\PearsonGroupwiseChem};
                \end{axis}
            \end{tikzpicture}
            \caption{\textbf{Chemical}}
            \label{subfig:PearsonCorr_chem_xgB}
        \end{subfigure}
        \begin{subfigure}[b]{0.245\linewidth}
        \begin{tikzpicture}[]
                \begin{axis}
                [
                    MediumBarPlot,
                    width = \linewidth,
                    height = 3.5cm,
                    ylabel = Pearson Correlation,
                    label style = {align = center, font = \footnotesize},
                    grid=major,
                    yticklabel=\pgfmathprintnumber{\tick},
                    xtick = data,
                    ymin = -1,ymax=1.01,
                    xmin = -0.45, xmax = 7.4,
                    xticklabels from table={\PearsonGroupwiseLM}{Feature},
                  x tick label style={font = \tiny,rotate = 90, anchor=east},
                    legend pos=north east,font = \tiny,
                  legend style={at={(0,1)}, anchor=south east}
                ]
                \addplot [PurpleBars] table [y={Pearson_Corr}, x expr=\coordindex] {\PearsonGroupwiseLM};
                \end{axis}
            \end{tikzpicture}
            \caption{\textbf{Landmine}}
            \label{subfig:PearsonCorr_lm_xgB}
        \end{subfigure}
        \begin{subfigure}[b]{0.245\linewidth}
        \begin{tikzpicture}[]
                \begin{axis}
                [
                    MediumBarPlot,
                    width = \linewidth,
                    height = 3.5cm,
                    ylabel = Pearson Correlation,
                    label style = {align = center, font = \footnotesize},
                    grid=major,
                    yticklabel=\pgfmathprintnumber{\tick},
                    xtick = data,
                    ymin = -1, ymax = 1.01,
                    xmin = -0.45, xmax = 7.4,
                    xticklabels from table={\PearsonGroupwisePK}{Feature},
                  x tick label style={font = \tiny,rotate = 90, anchor=east},
                    legend pos=north east,font = \tiny,
                  legend style={at={(0,1)}, anchor=south east}
                ]
                \addplot [PurpleBars] table [y={Pearson_Corr}, x expr=\coordindex] {\PearsonGroupwisePK};
                \end{axis}
            \end{tikzpicture}
            \caption{\textbf{Parkinson}}
            \label{subfig:PearsonCorr_pk_xgB}
        \end{subfigure}
        \vspace{-1em}
\caption{Pearson correlation between groupwise-task features and relative MTL gain for a group of tasks trained using \textbf{Extreme Gradient Boosting trees (XGBoost)}.}
\label{fig:pearsonCorr_tgp_xgB}
        \vspace{-0.5em}
\end{figure}

\cref{fig:pearsonCorr_tgp,fig:pearsonCorr_tgp_svm,fig:pearsonCorr_tgp_xgB} show the Pearson correlation coefficient between the groupwise-task features and relative MTL gains for task-groups with more than 2 tasks, trained using neural network, SVM, and extreme gradient boosting (XGBoost) trees, respectively. We describe the groupwise-task features in \cref{appendix:groupwise_feature_desc}. Over the four benchmarks and different MTL implementations, the mean of the pairwise MTL gains correlates the most with MTL gain for a group of tasks.

\subsection{Usefulness for Groupwise-Task Features}
\label{appendix:usefulness_group_svm}

\cref{fig:predictor_groupwise_svm} and \cref{fig:predictor_groupwise_xgb} show the usefulness of each groupwise-task feature for predicting relative MTL gain for groups of tasks trained using support vector machines (SVM) and extreme gradient boosting (XGBoost) trees, respectively. The features are described in \cref{appendix:groupwise_feature_desc}. 

\pgfplotstableread[col sep=comma]{data/Groupwise_Individual_Usefulness_School_avg_SVM.csv}\nnGroupwiseSch
\pgfplotstableread[col sep=comma]{data/Groupwise_Individual_Usefulness_Chemical_avg_SVM.csv}\nnGroupwiseChem
\pgfplotstableread[col sep=comma]{data/Groupwise_Individual_Usefulness_Landmine_avg_SVM.csv}\nnGroupwiseLM
\pgfplotstableread[col sep=comma]{data/Groupwise_Individual_Usefulness_Parkinsons_avg_SVM.csv}\nnGroupwisePK

\begin{figure}[h!]
    \begin{subfigure}[b]{0.245\linewidth}
    \begin{tikzpicture}[]
            \begin{axis}
            [
                MediumBarPlot,
                width = \linewidth,
                height = 3.5cm,
                ylabel = Usefulness (\textit{$R^2$}),
                label style = {align = center, font = \footnotesize},
                grid=major,
                yticklabel=\pgfmathprintnumber{\tick}{$\%$},,
                xtick = data,
                ymin = -2,
                xmin = -0.45, xmax = 7.8,
                xticklabels from table={\nnGroupwiseSch}{Feature},
              x tick label style={font = \tiny,rotate = 90, anchor=east},
                legend pos=north east,font = \tiny,
              legend style={at={(0,1)}, anchor=south east}
            ]
            \addplot [TealBars] table [y={Avg_R_SQUARE}, x expr=\coordindex] {\nnGroupwiseSch};
            \end{axis}
        \end{tikzpicture}
        \caption{\textbf{School}}
        \vspace{-0.5pt}
        \label{subfig:groupPredictor_sch_svm}
    \end{subfigure}
    \begin{subfigure}[b]{0.245\linewidth}
        \begin{tikzpicture}[]
                \begin{axis}
                [
                    MediumBarPlot,
                    width = \linewidth,
                    height = 3.5cm,
                    ylabel = Usefulness (\textit{$R^2$}),
                    label style = {align = center, font = \footnotesize},
                    grid=major,
                    yticklabel=\pgfmathprintnumber{\tick}{$\%$},,
                    xtick = data,
                    ymin = -2,
                    xmin = -0.45, xmax = 7.8,
                    xticklabels from table={\nnGroupwiseChem}{Feature},
                  x tick label style={font = \tiny,rotate = 90, anchor=east},
                    legend pos=north east,font = \tiny,
                  legend style={at={(0,1)}, anchor=south east}
                ]
                \addplot [TealBars] table [y={Avg_R_SQUARE}, x expr=\coordindex] {\nnGroupwiseChem};
                \end{axis}
            \end{tikzpicture}
            \caption{\textbf{Chemical}}
            \label{subfig:groupPredictor_chem_svm}
        \end{subfigure}
        \begin{subfigure}[b]{0.245\linewidth}
        \begin{tikzpicture}[]
                \begin{axis}
                [
                    MediumBarPlot,
                    width = \linewidth,
                    height = 3.5cm,
                    ylabel = Usefulness (\textit{$R^2$}),
                    label style = {align = center, font = \footnotesize},
                    grid=major,
                    yticklabel=\pgfmathprintnumber{\tick}{$\%$},,
                    xtick = data,
                    ymin = -2,
                    xmin = -0.45, xmax = 7.8,
                    xticklabels from table={\nnGroupwiseLM}{Feature},
                  x tick label style={font = \tiny,rotate = 90, anchor=east},
                    legend pos=north east,font = \tiny,
                  legend style={at={(0,1)}, anchor=south east}
                ]
                \addplot [TealBars] table [y={Avg_R_SQUARE}, x expr=\coordindex] {\nnGroupwiseLM};
                \end{axis}
            \end{tikzpicture}
            \caption{\textbf{Landmine}}
            \label{subfig:groupPredictor_lm_svm}
        \end{subfigure}
        \begin{subfigure}[b]{0.245\linewidth}
        \begin{tikzpicture}[]
                \begin{axis}
                [
                    MediumBarPlot,
                    width = \linewidth,
                    height = 3.5cm,
                    ylabel = Usefulness (\textit{$R^2$}),
                    label style = {align = center, font = \footnotesize},
                    grid=major,
                    yticklabel=\pgfmathprintnumber{\tick}{$\%$},
                    xtick = data,
                    ymin = -2,
                    xmin = -0.45, xmax = 7.8,
                    xticklabels from table={\nnGroupwisePK}{Feature},
                  x tick label style={font = \tiny,rotate = 90, anchor=east},
                    legend pos=north east,font = \tiny,
                  legend style={at={(0,1)}, anchor=south east}
                ]
                \addplot [TealBars] table [y={Avg_R_SQUARE}, x expr=\coordindex] {\nnGroupwisePK};
                \end{axis}
            \end{tikzpicture}
            \caption{\textbf{Parkinson}}
            \label{subfig:groupPredictor_pk_svm}
        \end{subfigure}
        \vspace{-1em}
\caption{Usefulness of features for predicting relative MTL gain when several tasks ($>2$) are trained together using \textbf{Support Vector Machines (SVM)}. The average pairwise relative MTL gain is the most effective in predicting relative MTL gain for a group for three benchmarks.}
\label{fig:predictor_groupwise_svm}
        \vspace{-0.5em}
\end{figure}
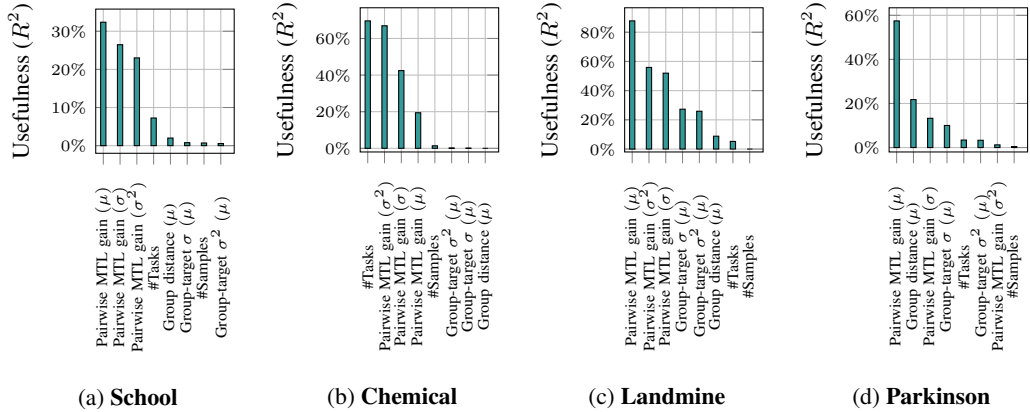
\pgfplotstableread[col sep=comma]{data/Groupwise_Individual_Usefulness_School_avg_xgBoost.csv}\nnGroupwiseSch
\pgfplotstableread[col sep=comma]{data/Groupwise_Individual_Usefulness_Chemical_avg_xgBoost.csv}\nnGroupwiseChem
\pgfplotstableread[col sep=comma]{data/Groupwise_Individual_Usefulness_Landmine_avg_xgBoost.csv}\nnGroupwiseLM
\pgfplotstableread[col sep=comma]{data/Groupwise_Individual_Usefulness_Parkinsons_avg_xgBoost.csv}\nnGroupwisePK

\begin{figure}[h!]
    \begin{subfigure}[b]{0.245\linewidth}
    \begin{tikzpicture}[]
            \begin{axis}
            [
                MediumBarPlot,
                width = \linewidth,
                height = 3.5cm,
                ylabel = Usefulness (\textit{$R^2$}),
                label style = {align = center, font = \footnotesize},
                grid=major,
                yticklabel=\pgfmathprintnumber{\tick}{$\%$},,
                xtick = data,
                ymin = -2,
                xmin = -0.45, xmax = 7.8,
                xticklabels from table={\nnGroupwiseSch}{Feature},
              x tick label style={font = \tiny,rotate = 90, anchor=east},
                legend pos=north east,font = \tiny,
              legend style={at={(0,1)}, anchor=south east}
            ]
            \addplot [TealBars] table [y={Avg_R_SQUARE}, x expr=\coordindex] {\nnGroupwiseSch};
            \end{axis}
        \end{tikzpicture}
        \caption{\textbf{School}}
        \vspace{-0.5pt}
        \label{subfig:groupPredictor_sch_xgB}
    \end{subfigure}
    \begin{subfigure}[b]{0.245\linewidth}
        \begin{tikzpicture}[]
                \begin{axis}
                [
                    MediumBarPlot,
                    width = \linewidth,
                    height = 3.5cm,
                    ylabel = Usefulness (\textit{$R^2$}),
                    label style = {align = center, font = \footnotesize},
                    grid=major,
                    yticklabel=\pgfmathprintnumber{\tick}{$\%$},,
                    xtick = data,
                    ymin = -2,
                    xmin = -0.45, xmax = 7.8,
                    xticklabels from table={\nnGroupwiseChem}{Feature},
                  x tick label style={font = \tiny,rotate = 90, anchor=east},
                    legend pos=north east,font = \tiny,
                  legend style={at={(0,1)}, anchor=south east}
                ]
                \addplot [TealBars] table [y={Avg_R_SQUARE}, x expr=\coordindex] {\nnGroupwiseChem};
                \end{axis}
            \end{tikzpicture}
            \caption{\textbf{Chemical}}
            \label{subfig:groupPredictor_chem_xgB}
        \end{subfigure}
        \begin{subfigure}[b]{0.245\linewidth}
        \begin{tikzpicture}[]
                \begin{axis}
                [
                    MediumBarPlot,
                    width = \linewidth,
                    height = 3.5cm,
                    ylabel = Usefulness (\textit{$R^2$}),
                    label style = {align = center, font = \footnotesize},
                    grid=major,
                    yticklabel=\pgfmathprintnumber{\tick}{$\%$},,
                    xtick = data,
                    ymin = -2,
                    xmin = -0.45, xmax = 7.8,
                    xticklabels from table={\nnGroupwiseLM}{Feature},
                  x tick label style={font = \tiny,rotate = 90, anchor=east},
                    legend pos=north east,font = \tiny,
                  legend style={at={(0,1)}, anchor=south east}
                ]
                \addplot [TealBars] table [y={Avg_R_SQUARE}, x expr=\coordindex] {\nnGroupwiseLM};
                \end{axis}
            \end{tikzpicture}
            \caption{\textbf{Landmine}}
            \label{subfig:groupPredictor_lm_xgB}
        \end{subfigure}
        \begin{subfigure}[b]{0.245\linewidth}
        \begin{tikzpicture}[]
                \begin{axis}
                [
                    MediumBarPlot,
                    width = \linewidth,
                    height = 3.5cm,
                    ylabel = Usefulness (\textit{$R^2$}),
                    label style = {align = center, font = \footnotesize},
                    grid=major,
                    yticklabel=\pgfmathprintnumber{\tick}{$\%$},
                    xtick = data,
                    ymin = -2,
                    xmin = -0.45, xmax = 7.8,
                    xticklabels from table={\nnGroupwisePK}{Feature},
                  x tick label style={font = \tiny,rotate = 90, anchor=east},
                    legend pos=north east,font = \tiny,
                  legend style={at={(0,1)}, anchor=south east}
                ]
                \addplot [TealBars] table [y={Avg_R_SQUARE}, x expr=\coordindex] {\nnGroupwisePK};
                \end{axis}
            \end{tikzpicture}
            \caption{\textbf{Parkinson}}
            \label{subfig:groupPredictor_pk_xgB}
        \end{subfigure}
        \vspace{-1em}
\caption{Usefulness of features for predicting relative MTL gain when several tasks ($>2$) are trained together using \textbf{Extreme Gradient Boost (XGBoost)} trees. Pairwise MTL gain ($\mu$) is the most useful groupwise-task feature.}
\label{fig:predictor_groupwise_xgb}
        \vspace{-1em}
\end{figure}
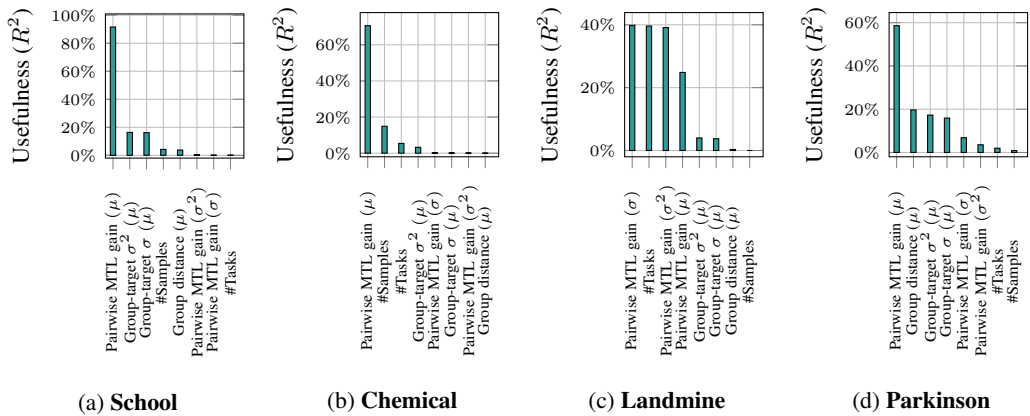

Similar to \cref{fig:predictor_groupwise}, we can see that the highest useful feature is the average of the pairwise MTL gains for all task pairs in a group. 
The number of tasks within a group, the average variance in the target attributes of the tasks
exhibit general usefulness for all benchmarks where MTL models were implemented with XGBoost trees (\cref{subfig:groupPredictor_sch_xgB}).

\subsection{Usefulness of Groupwise MTL-Gain Predictor}

In this section, we study the performance of our groupwise task-affinity predictor. By \emph{task-affinity}, we mean the relative improvement achieved through multi-task learning compared to individual single-task learning performance. We measure how much variance in the relative performance gain can be explained by our predictors ($R^2$-value). To obtain the architecture and input features of the groupwise MTL affinity predictor, we first conducted a search that combines hyper-parameter and feature selection. Utilizing the best architecture and groupwise-task feature set determined by this search, we train our final groupwise-MTL affinity predictor using the groups from our \textit{partition sample}, $\groupSet$, and evaluate on a hold-out test set.

\begin{table}[]
    \centering
    \caption{Usefulness of Groupwise MTL-Gain Predictor when MTL Models are Trained using \textbf{Neural Networks}}
    \label{tab:groupwise_affinitypredictor_rsq}
    \vspace{0.5em}
    \begin{tabular}{*{3}{|c}|}
    \hline
     
    & \textbf{Co-efficient of Determination ($R^2$)}
    & \textbf{Co-efficient of Determination ($R^2$)}\\
    
    \textbf{Dataset}
    & with affinity-predictor using
    & with affinity-predictor using \\

    & \textbf{avg. of Relative}
    & \textbf{Optimal Architecture}\\

    & \textbf{Pairiwise MTL Gain}
    & \textbf{and Input Features} (\cref{appendix:affinity_pred_architecture})\\ \hline

    \textbf{School} &  7.25\% & 16.73\%\\ \hline 
    \textbf{Chemical} &  43.35\% & 46.44\%\\ \hline 
    \textbf{Landmine} &  1.66\% & 4.91\%\\ \hline 
    \textbf{Parkinson} &  21.99\% & 29.56\%\\ \hline 
       
    \end{tabular}
    \vspace{-0.5em}
\end{table}

\cref{tab:groupwise_affinitypredictor_rsq} shows how much variance in the relative improvement in prediction performance of a group of tasks can be explained using a task-affinity predictor that takes only the average relative pairwise MTL gains as inputs, and with the final task-affinity predictor that takes the optimal set of groupwise-task features and architecture found from the combined hyper-parameter and feature-selection search. While \cref{tab:groupwise_affinitypredictor_rsq} shows results for groups of tasks trained using neural networks, \cref{tab:groupwise_affinitypredictor_rsq_SVM_xgb} show similar comparison for MTLs with support vector machines and XGBoost trees.

\begin{table}[h!]
    \centering
    \caption{Usefulness of Groupwise MTL-Gain Predictor when MTL Models are Trained using \textbf{Support Vector Machines (SVM)} and \textbf{Extreme Gradient Boosting (XGBoost)} Trees}
    \label{tab:groupwise_affinitypredictor_rsq_SVM_xgb}
    \resizebox{\linewidth}{!}{%
    \begin{tabular}{*{4}{|c}|}
    \hline
     
     \multirow{ 5}{*}{} &
    & \textbf{Co-efficient of Determination ($R^2$)}
    & \textbf{Co-efficient of Determination ($R^2$)}\\

    & \textbf{Dataset}
    & with affinity-predictor using
    & with affinity-predictor using  \\

    \textbf{Support}
    & & \textbf{avg. of Relative}
    & \textbf{Optimal Architecture}\\

\textbf{Vector}
   &  & \textbf{Pairwise MTL Gain}
    & \textbf{and Input Features} 
  (\cref{appendix:affinity_pred_architecture})\\ \cline{2-4}

\textbf{Machine}
    & \textbf{School}  & 32.37\% & 57.29\%\\ \cline{2-4} 
   &  \textbf{Chemical} & 19.42\% & 60.71\%\\ \cline{2-4}  
   &  \textbf{Landmine} & 87.77\% & 99.70\%\\ \cline{2-4}  
   &  \textbf{Parkinson}  & 57.47\% & 70.12\%\\   \hline \hline

     \multirow{ 5}{*}{} &
    & \textbf{Co-efficient of Determination ($R^2$)}
    & \textbf{Co-efficient of Determination ($R^2$)}\\

    & \textbf{Dataset}
    & with affinity-predictor using
    & with affinity-predictor using \\

    \textbf{Extreme}
    & & \textbf{avg. of Relative}
    & \textbf{Optimal Architecture}\\

\textbf{Gradient}
   &  & \textbf{Pairwise MTL Gain}
    & \textbf{and Input Features} 
  (\cref{appendix:affinity_pred_architecture})\\ \cline{2-4}

\textbf{Boosting}
   & \textbf{School}  &  91.56\% & 92.06\%\\ \cline{2-4} 
   &  \textbf{Chemical}  & 70.59\% & 79.56\%\\ \cline{2-4}  
   &  \textbf{Landmine} & 39.71\% & 49.02\%\\ \cline{2-4}  
   &  \textbf{Parkinson}  & 58.62\% & 80.32\%\\ \hline

    \end{tabular}
    }
\end{table}

The results reveal that the correct input features and a well-suited architecture can enable the groupwise MTL-gain predictor to effectively explain a significant proportion of the variance in relative pairwise MTL gain.

\section{Task Grouping Approach}

\subsection{Group Mutation}
\label{appendix:group_mutation}

Our randomized local search follows an iterative process: in each step, it generates a new partition $G'$ using a random mutation of the current partition $G$. 
In principle, this mutation selects one task at random and moves it from its current group to another randomly chosen group. However, this mutation must consider that removing and adding a task can destroy and create groups.
\begin{algorithm}[H]
\caption{\textbf{MutateGroups}$(G,\taskSet)$}
\label{alg:mutation}
\textbf{Input}: task set $\taskSet$, partition $G$

\begin{algorithmic}[1] %

\STATE  $t \gets $\textbf{RandomChoice}$(\taskSet)$
\STATE $g_\textit{old} \gets g \in G: \, t \in \taskSet_g$
    \IF {$|g_{\textit{old}}| = 1$}
        \STATE   $G' \gets G \setminus \{g_{\textit{old}}\}$
        \STATE   $g_\textit{new} \gets $ \textbf{RandomChoice}$\left(G'\right)$
        \STATE    $g_{\textit{new}}' \gets  g_{\textit{new}} \cup \{t\}$
    \ELSE
        \STATE $G' \gets G$ 
        \STATE $g_{\textit{old}}' \gets  g_{\textit{old}} \setminus \{ t \}$
        \STATE  $g_\textit{new} \gets$ \textbf{RandomChoice}$\left(G \setminus \{g_\textit{old}\} \cup \{\{t\}\}\right)$
        
        \IF{$g_\textit{new} = \{t\}$}
          \STATE $G' \gets G' \cup \{g_\textit{new}\}$
        \ELSE
          \STATE $g_{\textit{new}}' \gets  g_{\textit{new}} \cup \{t\}$
      \ENDIF
      
    \ENDIF

\STATE \textbf{return} $G'$
\end{algorithmic}
\end{algorithm}
\vspace{-1.75em}

We describe our random mutation approach in detail in \cref{alg:mutation}. The algorithm takes as input the current partition (i.e., set of groups) $G$ and the set of all tasks $\taskSet$.
First, it selects a task $t \in \taskSet$ uniformly at random from the set of all tasks (we let \textbf{RandomChoice} denote selecting an element uniformly at random from a set).

Then, we let $g_\textit{old}$ denote the group that contains task $t$ in the current partition $G$ (line 2).
If this group $g_\textit{old}$ contains only a single task (i.e., $|g_\textit{old}| = 1$), then the removal of task $t$ will empty the group.
In this case, we remove group $g_\textit{old}$ from the new partition $G'$ (line 4), we choose another group $g_\textit{new}$ from the remaining ones at random (line 5), and we add task $t$ to this new group $g_\textit{new}$ (line 6).
Otherwise, we do not remove group $g_\textit{old}$ from the new partition $G'$ (line 8); we just remove task $t$ from group $g_\textit{old}$ (line 9).
Finally, we choose another group $g_\textit{new}$, which can be either a new group $\{t\}$ or an existing group in the current partition $G$ (line 10).
In the first case, we add a new singleton group $\{t\}$ to the new partition $G'$ (line 12).
In the latter case, we add task $t$ to the new group $g_\textit{new}$ (line 14).

\subsection{Training and Query Time of Task-Affinity Predictor}
\label{appendix:queryTime}

In \cref{subsec:quick_reject}, we explain how we use the task-affinity predictor (i.e., relative MTL gain predictor)  before performing multi-task learning to speed up our randomized local search. We first train the task-affinity predictor on the $k$ initial random partitions. We also retrain the task-affinity predictor periodically during the search to improve its performance. The computational cost of the
initial training and periodic retraining of the task-affinity predictor is negligible compared to performing multi-task learning in each iteration of the search. 

\begin{table}[h!]
\centering
\caption{Training and Query Time of the Task-Affinity Predictor vs.  Training Time of MTL}
\vspace{0.5em}
\def\arraystretch{1.1}
\begin{tabular}{|c||c|c|c|}
\hline
\textbf{Benchmark} &  & \textbf{Task-Affinity} & \textbf{MTL} \\
\textbf{Dataset} & \textbf{Sample} & \textbf{Predictor Time} & \textbf{Training Time} \\
& & (minutes) & (minutes) \\ \hline\hline

\textbf{School}   & \makecell{2 groups \\$|g_1| = 13$\\$|g_2| = 15$} & 0.37                                      & 3.27                          \\\hline
\textbf{Chemical} & \makecell{2 groups \\$|g_1| = 8$\\$|g_2| = 9$}   & 0.11                                     & 7.04                          \\\hline
\textbf{Landmine} & \makecell{2 groups \\$|g_1| = 6$\\$|g_2| = 5$}  & 0.15                                     & 3.25            \\ \hline
\textbf{Parkinson} & \makecell{2 groups \\$|g_1| = 7$\\$|g_2| = 5$}  & 0.11                                     & 2.5            \\
\hline
\end{tabular}
\label{tab:query_time}
\end{table}

\cref{tab:query_time} reports the time that the task-affinity predictor takes to quickly reject unpromising mutations  and the time that it takes to train multi-task models for these mutations. The \emph{running time of the task-affinity predictor includes both the time to train the predictor on the initial partitions and the time to query the predictor} for two groups. For example, in the
School benchmark with two groups of 13 and 15 tasks, 
the total MTL training time (without quick reject) is more than 3.2 minutes, and the total time for training and querying the
task affinity predictor is only 0.37 minutes. For Chemical benchmark, the task-affinity predictor time (i.e., training and querying) is 0.11 minutes, while the MTL training time for two groups is 7.04 minutes.

\section{Task Grouping Results}
\label{appendix:extra_results}

\subsection{Task-Grouping Results and Comparison with Baselines - (SVM and XGBoost)}
In this subsection, we present the outcomes of our proposed approach along with a comparison to baselines utilizing a multi-task model employing a support vector machine (SVM) and extreme gradient boosting (XGBoost) trees. \cref{tab:comparison_with_SOTA_MTL_SVM} shows the results for task-grouping experiments when the underlying MTL models are implemented with support vector machines (SVM) and \cref{tab:comparison_with_SOTA_MTL_DT} shows similar comparisons with extreme gradient boosting (XGBoost) trees as MTL models.

\begin{table}[H]
    \centering
    \caption{Comparison of Different STL and MTL Models (using \textbf{Support vector Machines (SVM)}) and Grouping Approaches on Benchmark Datasets\newline\emph{Note that since exhaustive search is very far from being computationally feasible, we substitute it with a random search. Our approach consistently outperforms all baselines across all benchmark datasets.}}
    \label{tab:comparison_with_SOTA_MTL_SVM}
    \vspace{-0.025em}
    \setlength{\tabcolsep}{2.5pt}
    \resizebox{\linewidth}{!}{%
    \begin{tabular}{*{11}{|c}|}
    \hline
     &  & &  & \textbf{Pairwise} & \textbf{Pairwise} & \textbf{Simple} &\multicolumn{2}{c|}{\textbf{Clustering Algorithms}}&\textbf{Exhaustive} & {\textbf{Our Approach}}\\ \cline{8-9} 
     
    \textbf{Dataset} & \textbf{Tasks} & \textbf{Evaluation} & \textbf{STL} & \textbf{MTL} (all pairs) &  \textbf{MTL} (optimal) & \textbf{MTL} &\textbf{Hierarchical}& \textbf{$k$-Means}& (Random & (Random Search\\
    
     & \textbf{$n$} & \textbf{Metric} & $|g_i|=1$  & $|g_i|=2$  & $|g_i|=2$ & $|g_i|=n$ &(MTL Affinities)&(MTL Affinities)& Partitionings)& w. Quick Reject)\\ \hline \hline
    
    \textbf{School} &139 & $\sum$\textit{MSE} $\downarrow$ 
    & 108.75 & 107.50 & 99.91 
    & 115.58
    & 113.13 $\pm$ 0.1
    & 113.26 $\pm$ 0.06
    & 109.1 $\pm$ 0.0
    & \textbf{89.41 $\pm$ 0.72}\\ 
 
     \hline
    
    \textbf{Chemical} &35 &$\sum$\textit{log-loss} $\downarrow$ 
    & 16.82  & 24.62 & 1.40 
    &  19.92
    & 18.33 $\pm$ 0.05
    & 18.66 $\pm$ 0.18
    & 14.83 $\pm$ 0.0
    & \textbf{12.63 $\pm$ 0.15}
    
     \\ \hline
    
    \textbf{Landmine} & 29 &$\sum$\textit{log-loss} $\downarrow$ 
    & 38.73 & 6.72 & 3.98 
    & 6.74
    & 6.51 $\pm$ 0.34
    & 6.61 $\pm$ 0.34
    & 6.53 $\pm$ 0.0
    &  \textbf{4.82 $\pm$ 0.1} 
    
     \\\hline

    \textbf{Parkinson} &42 & $\sum$\textit{MSE} $\downarrow$ 
    & 266.73 & 318.84 & 128.74 
    & 357.60
    & 311.35 $\pm$ 0.27 
    & 319.95 $\pm$ 0.6 
    & 315.46 $\pm$ 0.0 
    & \textbf{197.76 $\pm$ 2.29}
    \\\hline

    \end{tabular}
    }
    \vspace{-1.5em}
\end{table}

\begin{table}[H]
    \centering
    \caption{Comparison of Different STL and MTL Models (using \textbf{Extreme gradient Boosting (XGBoost)} trees) and Grouping Approaches on Benchmark Datasets\newline\emph{Note that since exhaustive search is very far from being computationally feasible, we substitute it with a random search. Our approach consistently outperforms all baselines across all benchmark datasets.}}
    \label{tab:comparison_with_SOTA_MTL_DT}
    \vspace{-0.025em}
    \setlength{\tabcolsep}{2.5pt}
    \resizebox{\linewidth}{!}{%
    \begin{tabular}{*{11}{|c}|}
    \hline
     &  & &  & \textbf{Pairwise} & \textbf{Pairwise} & \textbf{Simple} &\multicolumn{2}{c|}{\textbf{Clustering Algorithms}}&\textbf{Exhaustive} & {\textbf{Our Approach}}\\ \cline{8-9} 
     
    \textbf{Dataset} & \textbf{Tasks} & \textbf{Evaluation} & \textbf{STL} & \textbf{MTL} (all pairs) &  \textbf{MTL} (optimal) & \textbf{MTL} &\textbf{Hierarchical}& \textbf{$k$-Means}& (Random & (Random Search\\
    
     & \textbf{$n$} & \textbf{Metric} & $|g_i|=1$  & $|g_i|=2$  & $|g_i|=2$ & $|g_i|=n$ &(MTL Affinities)&(MTL Affinities) & Partitionings)& w. Quick Reject)\\ \hline \hline
    
    \textbf{School} &139 & $\sum$\textit{MSE} $\downarrow$ 
    & 108.25 
    & 131.24
    & 52.38 
    & 130.10 $\pm$ 0.2
    & 126.98 $\pm$ 0.1
    & 128.48 $\pm$ 0.04
    & 126.04 $\pm$ 0.01
    & \textbf{104.86 $\pm$ 2.21}\\  
     \hline
    
    \textbf{Chemical} &35 &$\sum$\textit{log-loss} $\downarrow$ 
    & 17.22  & 15.66 & 14.04
    & 17.43 $\pm$ 0.1
    & 16.37 $\pm$ 0.1
    & 16.44 $\pm$ 0.32
    & 15.29 $\pm$ 0.0
    & \textbf{11.49 $\pm$ 0.16}\\ 
     \hline
    
    \textbf{Landmine} & 29 &$\sum$\textit{log-loss} $\downarrow$ 
    & 6.76 & 8.59 & 4.68 
    & 9.19 $\pm$ 0.6
    & 7.93 $\pm$ 0.46
    & 8.25 $\pm$ 0.49
    & 6.24 $\pm$ 0.0
    &\textbf{5.78 $\pm$ 0.14}\\ 
     \hline

     \textbf{Parkinson} &42 & $\sum$\textit{MSE} $\downarrow$ 
     & 305.15 
     & 152.09
     & 94.05
    &  246.75 $\pm$ 4.36
    & 99.76 $\pm$ 21.77
    & 105.88 $\pm$ 013.88
    & 79.04 $\pm$ 0.0
    & \textbf{24.71 $\pm$ 0.62} \\\hline

    \end{tabular}
    }
    \vspace{-1em}
\end{table}

In \cref{tab:comparison_with_SOTA_MTL_SVM,tab:comparison_with_SOTA_MTL_DT}, we report the performance of our proposed approach and compare it against the baselines. All the reported results for each approach have been gathered through multiple repetitions of MTL training and evaluation process. The average loss over all tasks after STL ($|g_i| = 1$) and pairwise MTL ($|g_i| = 2$) demonstrates the usefulness of MTL. Even when considering all possible pairings, rather than only those with the positive transfer of information during training, we observe substantial improvements. The results from multi-task learning (MTL) with all tasks trained together ($|g_i| = n$) demonstrate performance enhancements compared to single-task learning (STL). However, they also highlight that negative transfer of information can occur when dissimilar tasks are grouped together, leading to reduced performance compared to pairwise MTL.
For classical task-clustering algorithms, we report average losses over all the partitions in \cref{tab:comparison_with_SOTA_MTL_SVM,tab:comparison_with_SOTA_MTL_DT}, but the losses of the best partitions (reported in \cref{tab:best_partition_svm_dt}) are \emph{still significantly outperformed by our approach}.

\begin{table}[h!]
    \centering
    \caption{Losses for Best Partitions when MTL models are Implemented using SVM and XGBoost Trees.}
    \label{tab:best_partition_svm_dt}
    \vspace{0.5em}
    \def\arraystretch{1.1}
    \begin{tabular}{|c|c|c|c|}
    \hline
        \textbf{Dataset} & \textbf{Metric} & \textbf{SVM} & \textbf{XGBoost} \\ \hline
        School & $\sum$ \textit{MSE} & 113.13 & 126.98 \\\hline
        Chemical & $\sum$ \textit{log-loss} & 18.33 & 16.37 \\\hline
        Landmine & $\sum$ \textit{log-loss} & 6.51 & 7.93 \\\hline
        Parkinson & $\sum$ \textit{MSE} & 311.35 & 99.76 \\\hline
    \end{tabular}
    
\end{table}

\subsection{Clustering Algorithms}
Classic hierarchical clustering algorithms require us to quantify the distance between any two tasks.
We consider several approaches for quantifying distance; in \cref{tab:comparison_with_SOTA_MTL} and \cref{tab:comparison_with_SOTA_MTL_SVM,tab:comparison_with_SOTA_MTL_DT}, we report results for the best-performing approach.  
\cref{tab:clustering_results} shows results for all the approaches we use. 
All results are based on 10-fold cross-validation and gathered after performing the experiments multiple times.
The clustering approaches use the pairwise relative MTL gain as a distance metric. In order to ensure that the tasks with higher affinity have lower distances, we multiply the relative gain values by -1 before performing exponentiation.
We also consider applying a logistic function to ensure non-negativity. 

For MTLs with neural networks, another approach that we use for quantifying the distance between tasks is based on the dot products of task-specific network parameters ($\mathcal{W}_{t_i}\cdot\mathcal{W}_{t_j}$) from an MTL model trained on all tasks. We also use task-specific parameters as the vector representation of tasks for $k$-means clustering.
\begin{table}[h!]
    \centering
     \caption{Comparison of Clustering Approaches for different MTL models using \textbf{Neural Networks}, \textbf{Support Vector Machines (SVM)}, and \textbf{Extreme Gradient Boosting (XGBoost)} Trees.}
    \label{tab:clustering_results}
   \resizebox{\linewidth}{!}{%
    \begin{tabular}{*{7}{|c}|}
    \hline
    \multirow{ 8}{*}{\textbf{Neural Network}} & \multirow{ 4}{*}{\textbf{Dataset}} &  & \multicolumn{4}{c|}{\textbf{Clustering Approach}}\\ \cline{4-7}
    
     & &  \textbf{Evaluation} & \multicolumn{3}{c|}{\textbf{Hierarchical}} & \textbf{$k$-Means}\\ \cline{4-7}
    
    &  & \textbf{Metric} & \multicolumn{2}{c|}{\textbf{MTL Affinity}} & \multirow{ 2}{*}{\textbf{Parameter Vector}} & \multirow{ 2}{*}{\textbf{Parameter Vector}} \\ \cline{4-5}
    
    &  &  &  \textbf{Exponential} & \textbf{Logistic} &  & \\ \cline{2-7}
    
     & \textbf{School} & $\sum$MSE $\downarrow$ 
     & 103.78 $\pm$ 0.05 
     & 102.15 $\pm$ 0.58
     & 99.0 $\pm$ 0.0
     & 97.97 $\pm$ 0.0 \\ \cline{2-7}
   
   &   \textbf{Chemical} & $\sum$log-loss $\downarrow$ 
   & 17.28 $\pm$ 0.23 
   & 17.26 $\pm$ 0.15 
   & 17.12 $\pm$ 0.0
   & 17.5 $\pm$ 0.0\\ \cline{2-7}
  
  &     \textbf{Landmine} & $\sum$log-loss $\downarrow$ 
  & 5.32 $\pm$ 0.01 
  & 5.28 $\pm$ 0.08 
  & 5.25 $\pm$ 0.0 
  & 5.36 $\pm$ 0.0\\ \cline{2-7}
   
   &     \textbf{Parkinson} & $\sum$MSE $\downarrow$ 
   & 261.51 $\pm$ 2.73
   & 256.66 $\pm$ 1.35 
   & 265.21 $\pm$ 0.0 
   & 269.04 $\pm$ 0.0\\ \hline \hline

       & \multirow{ 4}{*}{\textbf{Dataset}} &  & \multicolumn{4}{c|}{\textbf{Clustering Approach}}\\ \cline{4-7}
    
     & &  \textbf{Evaluation} & \multicolumn{2}{c|}{\textbf{Hierarchical}} & \multicolumn{2}{c|}{\textbf{$k$-Means}}\\ \cline{4-7}
    
    &  & \textbf{Metric} & \multicolumn{4}{c|}{\textbf{MTL Affinity}} \\ \cline{4-7}
    
    \textbf{Support Vector}  &  &  &  \textbf{Exponential} & \textbf{Logistic} &  \textbf{Exponential} & \textbf{Logistic} \\ \cline{2-7}
    
    \textbf{Machines (SVM)} & \textbf{School} & $\sum$MSE $\downarrow$ 
     & 113.22 $\pm$ 0.1 
     & 113.13 $\pm$ 0.1 
     & 113.61 $\pm$ 0.2 
     & 113.26 $\pm$ 0.06 \\ \cline{2-7}
     
   &   \textbf{Chemical} & $\sum$log-loss $\downarrow$ 
   & 18.33 $\pm$ 0.05 
   & 18.48 $\pm$ 0.05  
   & 18.66 $\pm$ 0.18 
   & 19.0 $\pm$ 0.09\\ \cline{2-7}
   
  &     \textbf{Landmine} & $\sum$log-loss $\downarrow$ 
  & 6.64 $\pm$ 0.39 
  & 6.51 $\pm$ 0.34 
  & 6.61 $\pm$ 0.30
  &  6.64 $\pm$ 0.29\\ \cline{2-7}
  
   &     \textbf{Parkinson} & $\sum$MSE $\downarrow$ 
   & 311.35 $\pm$ 0.27 
   & 320.16 $\pm$ 0.53 
   & 319.95 $\pm$ 0.6 
   & 323.98 $\pm$ 0.42 \\ \hline \hline

      & \multirow{ 4}{*}{\textbf{Dataset}} &  & \multicolumn{4}{c|}{\textbf{Clustering Approach}}\\ \cline{4-7}
    
     & &  \textbf{Evaluation} & \multicolumn{2}{c|}{\textbf{Hierarchical}} & \multicolumn{2}{c|}{\textbf{$k$-Means}}\\ \cline{4-7}
    
    &  & \textbf{Metric} & \multicolumn{4}{c|}{\textbf{MTL Affinity}} \\ \cline{4-7}
    
    \textbf{Extreme Gradient} &  &  &  \textbf{Exponential} & \textbf{Logistic} &  \textbf{Exponential} & \textbf{Logistic} \\ \cline{2-7}
    
     \textbf{Boost (XGBoost)} & \textbf{School} & $\sum$MSE $\downarrow$ 
     & 127.21 $\pm$ 0.12 
     & 126.98 $\pm$ 0.1 
     & 128.65 $\pm$ 0.28 
     & 128.48 $\pm$ 0.19 \\ \cline{2-7}
     
   &   \textbf{Chemical} & $\sum$log-loss $\downarrow$ 
   & 16.37 $\pm$ 0.1  
   & 16.8 $\pm$ 0.1  
   & 17.21 $\pm$ 0.29 
   & 16.44 $\pm$ 0.32\\ \cline{2-7}
   
  &     \textbf{Landmine} & $\sum$log-loss $\downarrow$ 
  & 7.96 $\pm$ 0.25 
  & 7.93 $\pm$ 0.46 
  &  8.38 $\pm$ 0.37 
  & 8.25 $\pm$ 0.49 \\ \cline{2-7}
  
   &     \textbf{Parkinson} & $\sum$MSE $\downarrow$ 
   & 99.76 $\pm$ 21.77 
   &  102.17 $\pm$ 18.07 
   & 293.68 $\pm$ 67.26 
   &  105.88 $\pm$ 13.88 \\ \hline
    
    \end{tabular}
   }
   \end{table}

\clearpage

\end{document}